\documentclass{article}

\usepackage[preprint]{ProbNum25}
\usepackage[round]{natbib}

\usepackage[utf8]{inputenc} % allow utf-8 input
\usepackage[T1]{fontenc}    % use 8-bit T1 fonts
\usepackage{hyperref}       % hyperlinks
\usepackage{url}            % simple URL typesetting
\usepackage{booktabs}       % professional-quality tables
\usepackage{amsmath}
\usepackage{amsfonts}       % blackboard math symbols
\usepackage{nicefrac}       % compact symbols for 1/2, etc.
\usepackage{graphicx,wrapfig}
\usepackage{cancel}
\usepackage{microtype}      % microtypography
\usepackage{comment}
\usepackage{nccmath}
\usepackage{algpseudocode}

\usepackage[round]{natbib}

\DeclareMathOperator*{\argmin}{arg\,min}
\usepackage{amsthm}
\usepackage{tikz}
\usetikzlibrary{bayesnet}
\usetikzlibrary{arrows}
\usetikzlibrary{backgrounds}
\usepackage{algorithm}
\usepackage{algpseudocode}

\newcommand{\X}{\mathcal{X}} % smoothed input
\newcommand{\Xb}{\mathbf{X}} % smoothed input
\newcommand{\m}{\mathbf{m}} % smoothed input
\newcommand{\kb}{\mathbf{k}} % smoothed input
\newcommand{\mb}{\mathbf{m}} % smoothed input
\newcommand{\db}{\mathbf{d}} % smoothed input
\newcommand{\R}{\mathbf{R}} % smoothed input
\newcommand{\tb}{\mathbf{t}} % smoothed input
\newcommand{\x}{\mathbf{x}} % input point
\newcommand{\y}{\mathbf{y}} % input point
\newcommand{\mub}{\boldsymbol{\mu}} % pmf input
\newcommand{\Sigmab}{\boldsymbol{\Sigma}} % pmf input
\newcommand{\nub}{\boldsymbol{\nu}} % pmf input
\newcommand{\Pib}{\boldsymbol{\Pi}} % pmf input
\newcommand{\Gammab}{\boldsymbol{\Gamma}} % pmf input
\newcommand{\gammab}{\boldsymbol{\gamma}} % pmf input
\newcommand{\I}{\mathbf{I}} % pmf input
\newcommand{\A}{\mathbf{A}} % pmf input
\newcommand{\Taub}{\boldsymbol{\mathcal{T}}}

\makeatletter
\newcommand*{\centerfloat}{%
  \parindent \z@
  \leftskip \z@ \@plus 1fil \@minus \textwidth
  \rightskip\leftskip
  \parfillskip \z@skip}
\makeatother

\newtheorem{prop}{Proposition}
\newtheorem{corollary}{Corollary}

\probnumtitle{Natural Evolutionary Search meets Probabilistic Numerics}
\probnumauthors{%
\name{Pierre Osselin}%
\affiliation{Machine Learning Research Group, University of Oxford}
\and
\name{Masaki Adachi}%
\affiliation{Lattice Lab, Toyota Motor Corporation / Machine Learning Research Group, University of Oxford}
\and%
\name{Xiaowen Dong}%
\affiliation{Machine Learning Research Group, University of Oxford}
\and%
\name{Michael A. Osborne}%
\affiliation{Machine Learning Research Group, University of Oxford}
}

\probnumabstract{
Zeroth-order local optimisation algorithms are essential for solving real-valued black-box optimisation problems. Among these, Natural Evolution Strategies (NES) represent a prominent class, particularly well-suited for scenarios where prior distributions are available. By optimising the objective function in the space of search distributions, NES algorithms naturally integrate prior knowledge during initialisation, making them effective in settings such as semi-supervised learning and user-prior belief frameworks. However, due to their reliance on random sampling and Monte Carlo estimates, NES algorithms can suffer from limited sample efficiency. In this paper, we introduce a novel class of algorithms, termed Probabilistic Natural Evolutionary Strategy Algorithms (ProbNES), which enhance the NES framework with Bayesian quadrature. We show that ProbNES algorithms consistently outperforms their non-probabilistic counterparts as well as global sample efficient methods such as Bayesian Optimisation (BO) or $\pi$BO across a wide range of tasks, including benchmark test functions, data-driven optimisation tasks, user-informed hyperparameter tuning tasks and locomotion tasks.
}

\begin{document}

\section{Introduction}
From random search methods \citep{nelder1965simplex}, Genetic algorithms \citep{holland1992adaptation}, to Bayesian optimisation (BO; \citet{mockus1998application, osborne2009gaussian, garnett2023bayesian}) through metaheuristic algorithms \citep{kennedy1995particle}, both local and global zeroth order optimisation methods have been developed for decades and played an instrumental role in various fields such as machine learning \citep{bergstra2012random}, engineering design \citep{hansen2001completely} or finance \citep{cornuejols2018optimization}. In particular, natural evolutionary strategy algorithms (NES; \citet{wierstra2014natural}) is a popular subclass of local optimisation algorithms which adapts a search distribution to optimise a black-box function.

\begin{figure*}
    \centering
\includegraphics[width=\textwidth]{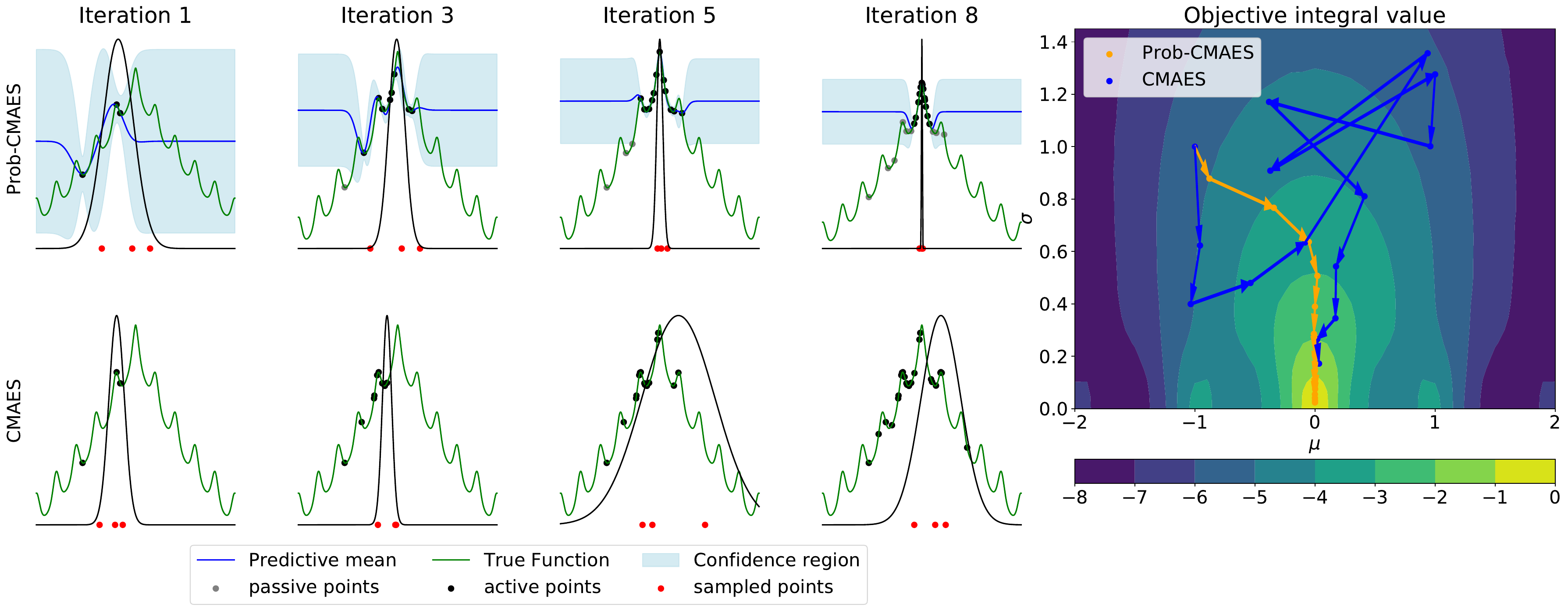}
    \vspace{-2em}
    \caption{Illustration of probabilistic CMAES}
    \label{fig:illustration}
\end{figure*}

More recently, attention has been drawn upon \textbf{user-informed} and \textbf{data-informed optimisation}. In the former case, information about promising points is given as input under the form of a probability distribution, forged through domain expertise of the user. The latter assumes an underlying data distribution exists which can (but not necessarily) inform us about optimal locations. This distribution can be given to us directly or from data samples $\{\x_i\}_{i \in \mathcal{D}}$. Examples of applications include hyperparameter tuning \citep{hvarfner2022pi}, where prior knowledge about the best parameters can be defined through past experiences, or chemical design tasks \citep{gomez2018automatic} where the objective is to generate molecules with certain properties through generative models trained with massive databases.

NES algorithms naturally exploits prior knowledge as they directly optimise in the space of search distribution via adapting its parameters to generate optimal points. In particular, CMAES has been shown to perform well under warm starts through transfer learning of previous experiments in the field of hyperparameter tuning \citep{nomura2021warm}. However, NES algorithms are memoryless and rely on random samples to gather knowledge of the objective function, which is less sample efficient than gathering data points in an active manner and can be detrimental in optimising black-box functions that are expensive to evaluate. On the other hand, BO algorithms are typically more sample efficient but do not easily integrate prior knowledge. 
This led to the development of multiple methods that usually design hand-made techniques such as modifying the BO acquisition function \citep{hvarfner2022pi}, or building a latent space with a particular training technique \citep{tripp2020sample}, metric \citep{grosnit2021high}, or regulariser \citep{maus2022local} on which regular BO is applied in a defined bounding box.

In this work, we make NES algorithms, which naturally handles prior knowledge, more sample efficient. To do this, we exploit the fact that they are a particular instantiation of Information-Geometric Optimisation Algorithms (IGO; \citet{ollivier2017information}) which perform natural gradient descent of the expectation of the objective function in the space of probability distributions. From this perspective, one can naturally use Bayesian Quadrature (BQ) to actively gather the most informative points and update distribution parameter based on Bayesian quadrature formula. 

Our contributions are summarised as follows:
\begin{enumerate}
    \item We describe the distribution of the gradient of the objective function integral under the Bayesian Quadrature and IGO framework (Section \ref{subsection:PNES})
    \item We derive the formula for the parameter updates for the search distribution of the probabilistic counterparts of XNES, SNES and rank-$\mu$ CMA-ES (Section \ref{subsection:paramupdate}) as well as implementation improvements (Section \ref{subsection:techimprov}).
    \item We showcase improved performances in the context of user and data informed optimisation on test functions (Section \ref{subsection:exp_testfunctions}), UCI datasets (Section \ref{subsection:exp_semisupervised}), optimisation in the latent space of generative models (Section \ref{subsection:exp_latentspace}) and hyperparameter tuning and locomotion tasks (Section \ref{subsection:exp_hyperparameter}).
    \item We perform ablation studies to analysis the robustness of our algorithms with respects to its parameters and validate the benefits of implementation choices (Section \ref{section:abl}).
    
\end{enumerate}

\section{Related Work}

\textbf{Prior-aware BO.} A class of method that aims at improving classical Gaussian Process (GP; \citet{stein1999interpolation, rasmussen2006gaussian}) based BO algorithms via incorporating prior knowledge \citep{bergstra2011algorithms, souza2021bayesian, jeong2021objective, nguyen2020knowing}. \citet{hvarfner2022pi, hvarfner2024a}, for example, uses decaying weights in the acquisition function of the GP model to influence initial selected points to be in regions of high likelihood of the prior distribution. \citet{adachi2022fast, adachi2024adaptive} aims at performing BO in the large batch setting via kernel recombination of a quadrature problem. To do this, this method keeps track of an approximated distribution of the objective function optimal points and, through this, it is possible to integrate a prior distribution.

\textbf{Latent space optimisation.} In the specific context of BO, some work has been carried out under the banner of `high dimensional BO' even though it could be cast as semi-supervised learning (SSL). In this line of work, a VAE \citep{kingma2014auto} is usually trained to model the training distribution of data and a standard BO with GP surrogate is performed in the latent space, no longer treated as the support of a random variable. Initial work \citep{tripp2020sample} improved this method by retraining the latent space by weighting important points in the loss, the weight being hand-crafted for the particular BO task. This work was later extended through adding a hand-crafted contrastive loss in the latent space, introducing some form of smoothness in the regression between the latent space and the output space \citep{grosnit2021high}. Recent work \citep{maus2022local} proposes to jointly train the latent space on unsupervised and supervised data points. In theory, the latent space should be constrained to perform well both for reconstruction and regression.

\textbf{Semi-supervised learning.} This setup is relatively close to SSL for regression, which is comparatively unexplored to SSL for classification. The differences are that in our setting, supervised points are gathered actively, and the goal is to optimize a function rather than modeling it. In this area, recent methods usually design a loss function taking the reconstruction error and regression error into account while imposing some `posterior regularization', which means hand crafting a term in the loss that would correspond to prior knowledge. Initial work \citep{jean2018semi} imposed some variance minimization constraints on the posterior over the unlabeled data, while later work \citep{xu2022semi} added a consistency regulariser, meaning that similar input points should have similar values.

\section{Background}
% \subsection{Setup}
% A decision making problem can be defined through a parameter space $\Theta$, an action space $\mathcal{A}$, and a loss function $l : \Theta \times \mathcal{A} \rightarrow \mathbb{R}$. Given data $\mathcal{D}$ the optimal decision (minimizing expected loss) is
% \begin{equation*}
%     a^* \in \argmin\limits_{a \in \mathcal{A}} \mathbb{E}_{\theta | \mathcal{D}}[l(\theta, a)]
% \end{equation*}
% In the example of BO, $\Theta = \mathcal{X}^{\mathcal{Y}}$ where $\mathcal{X}$ and $\mathcal{Y}$ are input and output spaces respectively, $\mathcal{A} = \mathcal{X}$ and $\theta | \mathcal{D}$ is usually defined as the posterior distribution of a GP model given gathered observations $\mathcal{D} = \{(\x_i, y_i)\}_{i \in [N]}$.

In this work, we will focus on optimization problems, although it could be applied for any statistical decision-making problems:
\begin{equation}\label{eq:prob_to_solve}
    \x^* \in \argmin\limits_{\x \in \mathcal{X}} f(\x).
\end{equation}
Furthermore, we assume we have a domain distribution $p(\x), \x \in \X$ informed via the user or from data. Next, we describe a natural framework to handle domain belief of where the optimal point resides.

\subsection{Natural Evolutionary Strategy \\ Algorithms (NES)}
Let a distribution $\nu_{\theta}$ with $\theta \in \Theta$ over the input space $\mathcal{X}$. Natural Evolution Strategies algorithms (NES) turn an initial optimisation problem in the input space $\mathcal{X}$ in parameter space $\Theta$ via:
\begin{equation} \label{eq:parame}
    \theta^* \in \argmin\limits_{\theta \in \Theta} \mathbb{E}_{\x \sim \nu_{\theta}}[f(\x)].
\end{equation}
From this formulation, it becomes feasible to leverage first order methods to optimise the Problem~(\ref{eq:parame}). Note that this formulation only requires a distribution over the input space which can be discrete, continuous, or mixed.
We have:
\begin{equation} \label{eq:gradient}
\nabla_{\theta} \int_X f(\x) \nu_{\theta}(d\x) = \int_X f(\x) \nabla_{\theta} \ln{\nu_{\theta}(\x)}  \nu_{\theta}(d\x),
\end{equation}
where the right-hand side can be estimated through Monte Carlo samples from $\nu_{\theta}$.
The success of NES algorithms also relies on a set of techniques described in \citet{wierstra2014natural}. In particular, the vanilla gradient is not scale invariant and leads to gradient explosion when the variance of the distribution becomes small. Instead, the \textit{natural gradient} $\widetilde{\nabla}_{\theta} = \mathbf{F}^{-1} \nabla_{\theta}$ where $\mathbf{F}$ is the \textit{Fisher Information Matrix} (see more details in the Appendix~\ref{appendix:naturalgradient}) has been shown to be scale-invariant and more effective. Additionally, empirical studies have shown using a fitness shaping function $W_{\theta}^{f}(x)$ instead of the raw function $f$ provides algorithmic robustness. The shaping function $W_{\theta}^{f}(x)$ is taken as the first $q^{th}$ quantile function.  This family of algorithm has also been cast in the framework of information-geometric optimisation with a continuous-time gradient flow in the parameter space \citep{ollivier2017information}.

Canonical NES algorithms consist in \textbf{1)} drawing $\lambda$ candidates $\mathbf{z}_i$ from the distribution $\nu_{\theta}$ \textbf{2)} evaluate the objective functions $f(\mathbf{z}_i)$ \textbf{3)} Estimate the natural gradient of the objective and \textbf{4)} update the parameters of the distribution.

\subsection{Gaussian Process model}

GPs are the most popular surrogate models to perform BO. This class of model gives analytic formula for its posterior distribution over test points to evaluate. Formally, a GP is denoted as $f \sim \mathcal{GP}(\mb, \kb)$, where $\mb (\x)$ and $\kb\left({\x},{\x}'\right)$ are the mean function and the covariance function (or the \textit{kernel}), respectively. While the mean function is often set to zero or a constant function, the covariance function encodes our belief on the property of the function we are modeling, such as its smoothness and periodicity. The covariance function typically has some kernel hyperparameters $\theta$ and are optimised by maximising the \textit{log-marginal likelihood}. Given dataset $\mathcal{D} = \left\{ \left( \x_i, y_i \right)_{i \in [N]}\right\}$, we assume $y_i = f(x_i) + \xi_i$, where $\xi_i$ is i.i.d. $\sigma_\mathrm{noise}$-sub-Gaussian noise with fixed $\sigma_\mathrm{noise} > 0$.
With $\mb(\cdot)$ and $\kb(\cdot, \cdot)$ defined, a GP gives analytic posterior mean 
%$\mu(\x_{*} \vert \mathcal{D}) = \mb(\x_{*}) + \mathbf{k}(\x_{*}, \mathbf{X}) \mathbf{K}^{-1}_{XX}\left(\mathbf{y} - \mb\left(\Xb\right) \right)$
$\mathbf{m}_\mathrm{post}(\x_{*}) = \mb(\x_{*}) + \mathbf{k}(\x_{*}, \mathbf{X}) \left[\kb_{XX} + \sigma_\mathrm{noise}^{2}\mathbf{I}\right]^{-1} \left(\mathbf{y} - \mb\left(\Xb\right) \right)$
and variance 
%$\kb(\x, \x' \vert \mathcal{D}) = \kb(\x, \x') - \mathbf{k}(\x, \mathbf{X}) \mathbf{K}^{-1}_{XX}\mathbf{k}(\mathbf{X}, \x'))$
$\kb_\mathrm{post}(\x, \x') = \kb(\x, \x') - \mathbf{k}(\x, \mathbf{X}) \left[\kb_{XX} + \sigma_\mathrm{noise}^{2}\mathbf{I}\right]^{-1}$ $\mathbf{k}(\mathbf{X}, \x')$
estimates on an unseen configuration $\x_{*}$, where $[\kb_{XX}]_{i,j}=\kb(\x_i, \x_j)$ is the $(i, j)$-th element of the Gram matrix induced on the $(i, j)$-th training samples by $\kb(\cdot, \cdot)$, the covariance function, and $\textbf{I}$ is an identity matrix.

\subsection{Bayesian Quadrature} \label{background:BQ}

Bayesian Quadrature (BQ; \citet{o1991bayes, rasmussen2003bayesian, hennig2022probabilistic}) is a framework leveraging Bayesian functional modelling to compute integral efficiently along its value uncertainty. If:
\begin{equation*}
    F = \int_{\mathcal{X}}f(\x)d\nu(\x) \quad f \sim \mathcal{GP}(\mathbf{m}, \mathbf{k}),
\end{equation*}

then:
\begin{equation*}
    F \mid \mathcal{D} \sim \mathcal{N}(F; m, v), %\text{ where} \begin{cases} 
  %\end{cases}
\end{equation*} where
\begin{equation*}
\begin{split}
    m &= \m_0 + \tb_X^{T}\kb_{XX}^{-1}(\y - \mathbf{m}_X),\\
    v &= \mathbf{R} - \tb_X^{T}\kb_{XX}^{-1}\tb_X,\\
    \m_0 &= \int_{\X} \m(\x) d\nu(\x), \\
    \tb(\x_i) &= \int_{\X} \kb(\x, \x_i)d\nu(\x), \\
    \mathbf{R} &= \iint_{\X} \mathbf{k}(\x, \x')d\nu(\x)d\nu(\x'). \\
\end{split}
\end{equation*}
For certain distribution $\nu$ and kernel $\kb$ closed form solution are available for the previous formula (see \ref{appendix:formula}). 
BQ allows to select evaluation nodes for the integral by maximising some acquisition function, one classical and common rule is the variance reduction acquisition function which targets the variance of the integrand:
\begin{equation*}
\begin{aligned}
    a_\mathrm{VR}(\mathbf{x}_{*}) =& \frac{\Delta\mathbb{V}[Z; \mathbf{x}_{*}]}{\mathbb{V}[Z \mid \mathcal{D}]}\\
    =& \frac{\mathbb{V}[Z \mid \mathcal{D}] - \mathbb{V}[Z \mid \mathcal{D} \cup (\mathbf{x}_{*}, \mathbf{y}_{*}) ] }{\mathbb{V}[Z \mid \mathcal{D}]}.
\end{aligned}
\end{equation*}

\section{Probabilistic NES}\label{subsection:PNES}

We model $f \sim \mathcal{GP}(\mb, \kb)$ as a GP and denote $g(\theta) := \mathbb{E}_{x \sim \nu_{\theta}}[f(x)]$. Given a surrogate model of the objective, we can derive a probabilistic version of all the steps taken in the canonical NES algorithm. In particular, candidates $\mathbf{z}_i$ can be actively generated to maximally inform about the integral and the natural gradient can be estimated via a BQ rules. The final algorithm is given in Algorithm~\ref{alg:main_alg}.

\subsection{Gradient Distribution via Bayesian \\ Quadrature Rules} 
In order to fit BQ within the NES framework, we study the distribution of the gradient $\nabla_{\theta} g(\theta) \mid \mathcal{D}$:
\begin{equation*}
\begin{split}
    \nabla_{\theta} g(\theta) \mid \mathcal{D} = \int_{\X} \nabla_{\theta} \ln{\nu_{\theta}}(\x) f(\x)d\nu_{\theta}(\x),
\end{split}
\end{equation*}
with 
\begin{equation*}
    f \mid \mathcal{D} \sim \mathcal{GP}(\mathbf{m}_\mathrm{post}, \kb_\mathrm{post}).
\end{equation*}

\begin{prop}[\textbf{Gradient via BQ}]\label{prop:gradient}
Let $\otimes$ be the outer product operator. The random vector $\nabla_{\theta} g(\theta) \mid \mathcal{D}$ is a multi output GP over the parameter space $\Theta$, and the probabilistic structure of $g(\theta)$ and $\nabla_{\theta} g(\theta)$ are the following:

\begin{equation} \label{eq:covariancestructure}
\begin{split}
    \mathbb{E}(g(\theta)) &= \int_{\X} \mathbf{m}_\mathrm{post}(\x)d\nu_{\theta}(\x)\\
    &= \m + \mathbf{t}_{kX}^{T}\left[\kb_{XX} + \sigma_\mathrm{noise}^{2}\mathbf{I}\right]^{-1} \left(\mathbf{Y} - \mathbf{m}(\Xb) \right),\\ \mathbb{E}(\nabla_{\theta} g(\theta)) &= \int_{\X} \nabla_{\theta} \ln{\nu_{\theta}}(\x) \mathbf{m}_\mathrm{post}(\x)d\nu_{\theta}(\x)\\
    &= \boldsymbol{\mathcal{M}} + \boldsymbol{\mathcal{T}}_{kX}^{T}\left[\kb_{XX} + \sigma_\mathrm{noise}^{2}\mathbf{I}\right]^{-1} \left(\mathbf{Y} - \mathbf{m}(\Xb) \right),\\
    \mathrm{Cov}(g(\theta_1),&\,\, g(\theta_2)) = \iint_{\X} \kb_\mathrm{post}(\x, \x')d\nu_{\theta_1}(\x)d\nu_{\theta_2}(\x') \\
    &= \mathbf{R}_{12} - \mathbf{t}_{1X}^{T}\left[\kb_{XX} + \sigma_\mathrm{noise}^{2}\mathbf{I}\right]^{-1}\mathbf{t}_{2X},\\
    \mathrm{Cov}(g(\theta_1), & \,\, \nabla_{\theta} g(\theta_2)) \\
    &= \iint_{\X} \nabla_{\theta} \ln{\nu_{\theta_{2}}}(\x') \kb_\mathrm{post}(\x, \x') d\nu_{\theta_1}(\x)d\nu_{\theta_2}(\x') \\
    &= \mathbf{R}_{12'} - \mathbf{t}_{1X}^{T}\left[\kb_{XX} + \sigma_\mathrm{noise}^{2}\mathbf{I}\right]^{-1} \boldsymbol{\mathcal{T}}_{2X},\\
    \mathrm{Cov}(\nabla_{\theta} g(&\theta_1),  \nabla_{\theta} g(\theta_2)) \\
    &= \iint_{\X} \nabla_{\theta} \ln{\nu_{\theta_1}}(\x) \otimes \nabla_{\theta} \ln{\nu_{\theta_2}}(\x') \kb_\mathrm{post}(\x, \x') \\ 
 & \hspace{135pt} d\nu_{\theta_1}(\x)d\nu_{\theta_2}(\x') \\
    &= \mathbf{R}_{1'2'} - \boldsymbol{\mathcal{T}}_{1X}^{T}\left[\kb_{XX} + \sigma_\mathrm{noise}^{2}\mathbf{I}\right]^{-1}\boldsymbol{\mathcal{T}}_{2X},
\end{split}
\end{equation}
with
\begin{subequations}
\begin{align*}
    \m &= \int_{\X} \m(\x) d\nu_{\theta}(\x), \\
    \boldsymbol{\mathcal{M}} &= \int_{\X} \nabla_{\theta} \ln{\nu_{\theta}}(\x) \m(\x) d\nu_{\theta}(\x), \\
    \mathbf{R}_{12} &= \iint_{\X} \kb(\x, \x')d\nu_{\theta_1}(x)d\nu_{\theta_2}(\x'), \\
    \mathbf{R}_{12'} &= \iint_{\X} \nabla_{\theta} \ln{\nu_{\theta_2}}(\x') \kb(\x, \x') d\nu_{\theta_1}(\x)d\nu_{\theta_2}(\x'),\\
    \mathbf{R}_{1'2'} &= \iint_{\X} \nabla_{\theta} \ln{\nu_{\theta_1}}(\x) \otimes \nabla_{\theta} \ln{\nu_{\theta_2}}(\x')\kb(\x, \x')\\ & \hspace{125pt}d\nu_{\theta_1}(\x)d\nu_{\theta_2}(\x'),\\
    \left[\mathbf{t}_{kX}\right]_{i} &= \int_{\X} \kb(\x, \x_i)d\nu_{\theta_k}(\x), \\
    \left[\boldsymbol{\mathcal{T}}_{kX}\right]_{ij} &= \int_{\X} \left(\nabla_{\theta} \ln{\nu_{\theta_k}}\right)_{j}(\x) \kb(\x, \x_i)d\nu_{\theta_k}(\x).
\end{align*}
\end{subequations}
\end{prop}

\begin{corollary}[\textbf{Analytical gradient}]\label{cor:analytical}
    Let $\nu_\theta(\textbf{x})$ be a multivariate normal distribution $\mathcal{N}(\textbf{x}; \boldsymbol{\mu}, \boldsymbol{\Sigma})$, 
    and the kernel be radial basis function (RBF) $k(\textbf{x}, \textbf{x}^\prime) = \zeta \mathcal{N}(\textbf{x}; \textbf{x}^\prime, \Lambda)$. All integration in Proposition~\ref{prop:gradient} becomes closed-form.
\end{corollary}
See Appendix \ref{appendix:formula} for the detailed proof and relevant notations. As demonstrated in Corollary~\ref{cor:analytical}, we derive the closed-form expressions for the gradient and its covariance. This paper primarily focuses on the combination of a Gaussian distribution with an RBF kernel. A comprehensive list of closed-form combinations can be found in Table 1 of \citet{briol2019probabilistic}.

This description of the random variable $\nabla_{\theta} g(\theta)$ allows us to select a gradient direction in the space of parameters $\Theta$, and informed by a GP surrogate model. In particular, it is possible to either choose the \textbf{expected} gradient or a \textbf{sampled} gradient. We explore the benefits of both in our experiments through an ablation study in Figure~\ref{fig:abl_sampled}.

\subsection{Parameter Updates of Probabilistic Evolutionary Strategy Algorithms} \label{subsection:paramupdate}
The most common search distribution used for evolutionary strategy algorithm is the multivariate normal distribution for which many algorithms has been developed, including CMAES, XNES or SNES. We will assume an input distribution $\nu(\x) = \mathcal{N}(\x; \mub, \Sigmab)$, a surrogate model $f \sim \mathcal{GP}(\mb, \kb)$, and multivariate normal kernel $\kb(\x, \x') = \theta \mathcal{N}(\x; \x', \Lambda)$. We suppose we gathered the dataset $\mathcal{D} = \left\{ \left( \x_i, y_i \right)_{i \in [N]}\right\}$.

In the NES framework, the parameters are updating via performing one step of a natural gradient descent with a Monte Carlo estimation of the gradient. In this section, we provide the probabilistic equivalent of these updates via taking the \textbf{expected} gradient estimate through our BQ rule. We derive these updates for the Rank-$\mathbf{\mu}$ CMA-ES algorithm below and for XNES and SNES algorithms in Appendix \ref{appendix:probes}.

\textbf{Rank-$\mathbf{\mu}$ CMA-ES.} 
Let $n$ be the size of the population generated at each generation, the Rank-$\mu$ updates of CMA-ES are:
\begin{equation*}
\begin{split}
    \mub^{t+1} &= \mub^{t} + \eta_{m}\sum\limits_{i=1}^{n} w_i \left( \x_i - \mub^t\right), \\
    \Sigmab^{t+1} &= \Sigmab^{t} + \eta_{c}\sum\limits_{i=1}^{n} w_i \left( \left( \x_i - \mub^t \right) \left( \x_i - \mub^t \right)^{T} - \Sigmab^{t}\right),
\end{split}
\end{equation*}
where $w_i$ are weights obtained from the ranked values of the objective function $f$. It has been shown that this update corresponds to a natural gradient update with $\eta_{m} = \eta_{c}$ \citep{akimoto2010bidirectional}.
\begin{prop}[\textbf{Probabilistic CMA-ES}]
Let $\odot$ be Hadamard product. The update for the probabilistic Rank-$\mu$ CMA-ES is:
\begin{equation*}
\begin{split}
    \mub^{t+1} &= \mub^{t} + \eta \Sigmab\left(\Sigmab^{t} + \Lambda\right)^{-1}\sum\limits_{i=1}^{N} w_i \left( \x_i - \mub^t\right), \\
    \Sigmab^{t+1} &= \Sigmab^{t} + \eta \Sigmab^{t}\left(\Sigmab^{t} + \Lambda\right)^{-1} \Bigg[\sum\limits_{i=1}^{N} w_i \Big(\\
    & \left( \x_i - \mub^t \right) \left( \x_i - \mub^t \right)^{T} - \left(\Sigmab^{t} + \Lambda\right)\Big) \Bigg] \left(\Sigmab^{t} + \Lambda\right)^{-1}\Sigmab^{t} \\
    \mathbf{w} &= \left[\kb_{\Xb\Xb} + \sigma_\mathrm{noise}^{2}\mathbf{I}\right]^{-1} (\y  - \mb_{\Xb}) \odot \mathcal{N}\left(\Xb; \mub, \Lambda + \Sigmab^{t}\right)
\end{split}
\end{equation*}
\end{prop}
\begin{proof}
We perform a natural gradient descent step with a learning step $\eta$. Given the parameterisation $\mathbf{\Theta} = \left(\mub, \Sigmab\right)$, the computation of the gradient of the expectation of the integral is given in Appendix~\ref{appendix:probaline} and the Fischer information matrix is given in Appendix~\ref{appendix:naturalgradient}.
\end{proof}

\subsection{Implementation Choices}\label{subsection:techimprov}
\begin{figure*}[h]
    \centering
\includegraphics[width=\textwidth]{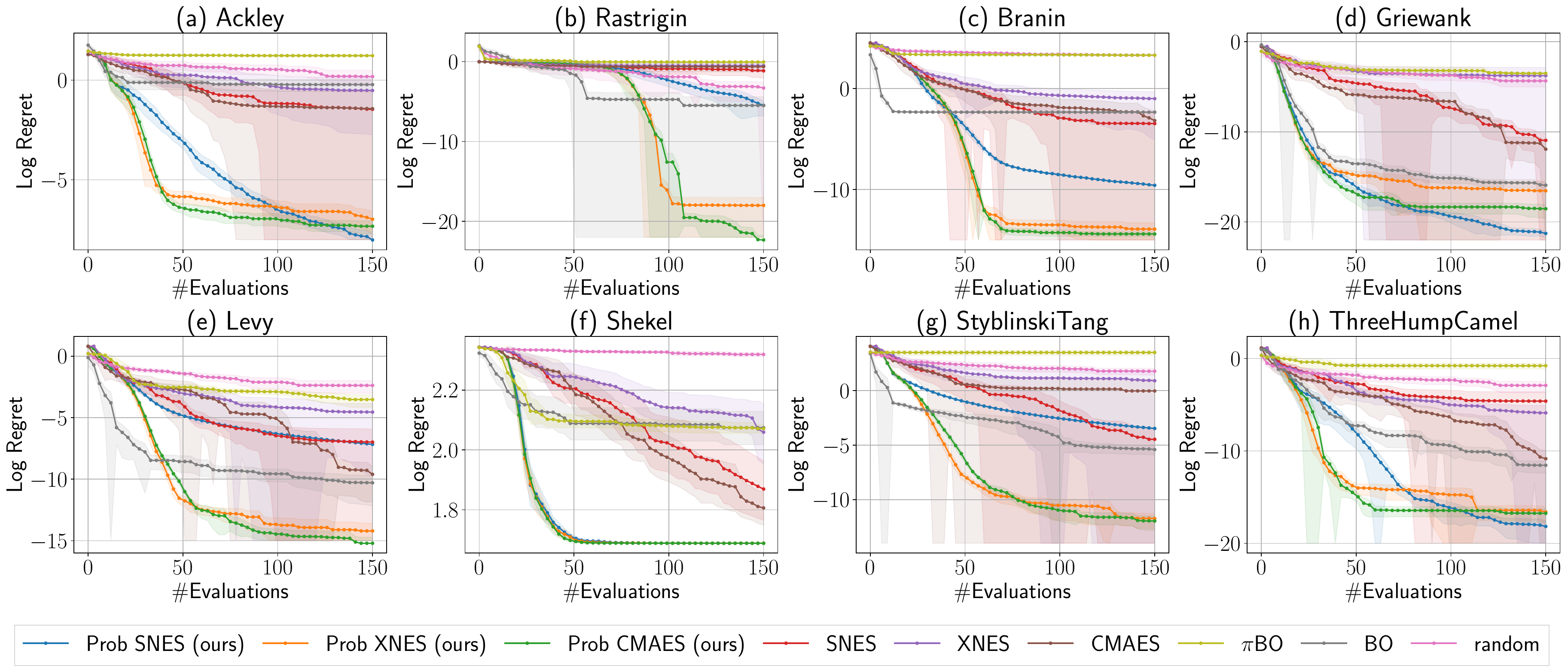}
    \vspace{-2em}
    \caption{Test Functions optimisation}
    \label{fig:test_function}
\end{figure*}

\begin{figure}[t] 
\begin{algorithm}[H]
\begin{footnotesize}
\caption{ProbNES algorithm} \label{alg:main_alg}
\begin{algorithmic}[1]
\State \textbf{inputs}: Number of random points at initialisation $N_0$, step size $\mu$, total number of iterations $T$, batch size $N$, initial prior distribution $\nu_{\theta_0} \in p(\mathcal{X})$ parameterised by $\theta_0 \in \Theta$, Gaussian Process (GP) kernel $\kb$ and objective function to optimise $f$.
\State \textbf{output}: Minimiser $x^*$ of the function $f(x), x \in \mathcal{X}$. 
\State \textbf{initialise} Initialise data $\mathcal{D} \leftarrow \{\}$, initialise GP surrogate model. 

%\For {$t = 0, ..., T - 1$}
\State \textbf{for} $t = 0, ..., T - 1$ \textbf{do}
    \State $\quad$ Select $N$ points $\{x_i\}_i$ actively via a batch BQ.
    \label{alg_line:batch_BQ}
    \State $\quad$ evaluate $y_i = f(x_i)$.
    \State $\quad$ Update $\mathcal{D} \leftarrow \mathcal{D} \cup \{(x_i, y_i)\}_{i=1:N}$
    \State $\quad$ Update GP on a subset of active points $\Tilde{\mathcal{D}}$.
    \label{alg_line:local_model}
    \State $\quad$ Compute the expected gradient $\db_t = \mathbb{E}\left[\nabla_{\theta} g(\theta_t)\right]$.
    \label{alg_line:gradient}
    \State $\quad$ Compute the Fisher Information Matrix $\boldsymbol{F}(\theta_t)$
    \State $\quad$ Update $\theta_{t+1} \leftarrow \theta_{t} + \mu \boldsymbol{F}^{-1}(\theta_t)\db_t $
\State \textbf{end for}

\State \textbf{return} Point $x^*$ that optimises $f(.)$ from all observations.
\end{algorithmic}
\end{footnotesize}
\end{algorithm}
\vspace{-2em}
\end{figure}

We introduce our algorithm, \emph{ProbNES}, in Algorithm~\ref{alg:main_alg}, accompanied by a detailed, line-by-line explanation of the implementation choices.

\textbf{Challenge: non-stationarity.} 
GPs assume the black-box function $f$ was sampled from a single prior distribution with, given a standard kernel (e.g., RBF kernel), a uniform level of regularity. In situation of non-stationary functions, GPs can fail to model appropriately and results in lengthscales that are underestimated. In ProbNES, this can lead to mis-calibration of the gradient estimates and suboptimal queried points. On the other hand, original NES algorithms does not suffer from such problem as they are scale invariant \citep{ollivier2017information}. Among the various approaches to handling non-stationarity (e.g., input warping \citet{snoek2014input}), we adopt a local modeling strategy for computational efficiency and closed-form expression (c.f., input warping requires another intractable integral approximation). By restricting the domain to a sufficiently small region defined by search distribution $\nu_\theta(\textbf{x})$, this area can be approximately viewed as stationary. \citet{eriksson2019scalable} empirically demonstrate the effectiveness of this approach, which has been gaining popularity in the BO literature (c.f., they employed hyperrectangle local region).

\textbf{Line~\ref{alg_line:batch_BQ}: Active sample selection.} 
With BQ, we are able to actively select the next points to query. To do so, we leverage the variance reduction acquisition function described in Section~\ref{background:BQ}. Formally,
\begin{align*}
    \textbf{x}_t = \arg \max_{\textbf{x} \in \mathcal{X}^g} a_\text{VR}(\textbf{x}),
\end{align*}
where $\textbf{x}_t = \{x_i\}_{i=1}^N$ are the batch of next queries, $\mathcal{X}^g$ is the `local' domain to explore. 
We define the local domain as $\mathcal{X}^g = \{ x \in \mathcal{X} \mid d_{\mub, \Sigmab}(\x) \leq \chi^2_{\alpha, 1-d}\}$, where $d_{\mub, \Sigmab}(\x) = \sqrt{\left(\x - \mub\right)^{\top}\Sigmab^{-1}\left(\x - \mub\right)}$ is the \textbf{Mahalanobis} distance, and $\chi^{2}_{\alpha, 1-d}$ is a threshold, the Chi-squared $1-\alpha$ critical value  with $d$ degree of freedom, with $\alpha = 99.73 \%$, inspired by \citet{ngo2024high}. This can be somewhat compared to the `$3\sigma$ rule' in normal distributions.

For the ablation study, we compared against the following cases: \textbf{1)} random sampling $\textbf{x}_t \sim \nu_\theta(\textbf{x})$, \textbf{2)} `best' random sample according to the acquisition function, \textbf{3)} `local' optimisation $\mathcal{X}^g$, \textbf{4)} `global' search $\mathcal{X}^g = \mathcal{X}$. 

\textbf{Line~\ref{alg_line:local_model}: Local modelling.} 
We also apply local domain approach to surrogate modelling. We consider \textbf{active} dataset, $\Tilde{\mathcal{D}} := \{(x_k, y_k)_{k=1}^{NT} \in \mathcal{D} \mid x_k \in \mathcal{X}^g\}$ (and \textbf{passive} dataset $\Bar{\mathcal{D}} := \mathcal{D} \textbackslash \Tilde{\mathcal{D}}$). We condition the GP only on active dataset $\Tilde{\mathcal{D}}$ for stationarity.

\textbf{Line~\ref{alg_line:gradient}: Analytical gradient.} BQ can offer the closed-form expected gradient $\db_t = \mathbb{E}\left[\nabla_{\theta} g(\theta_t)\right]$ (see Corollary~\ref{cor:analytical}). To understand the efficacy of BQ, we compared with the sampled gradient $\db_t \sim \nabla_{\theta} g(\theta_t)$, a rough estimation.

\section{Experiments} \label{subsection:exp}
\begin{figure*}[h]
    \centering
\includegraphics[width=\textwidth]{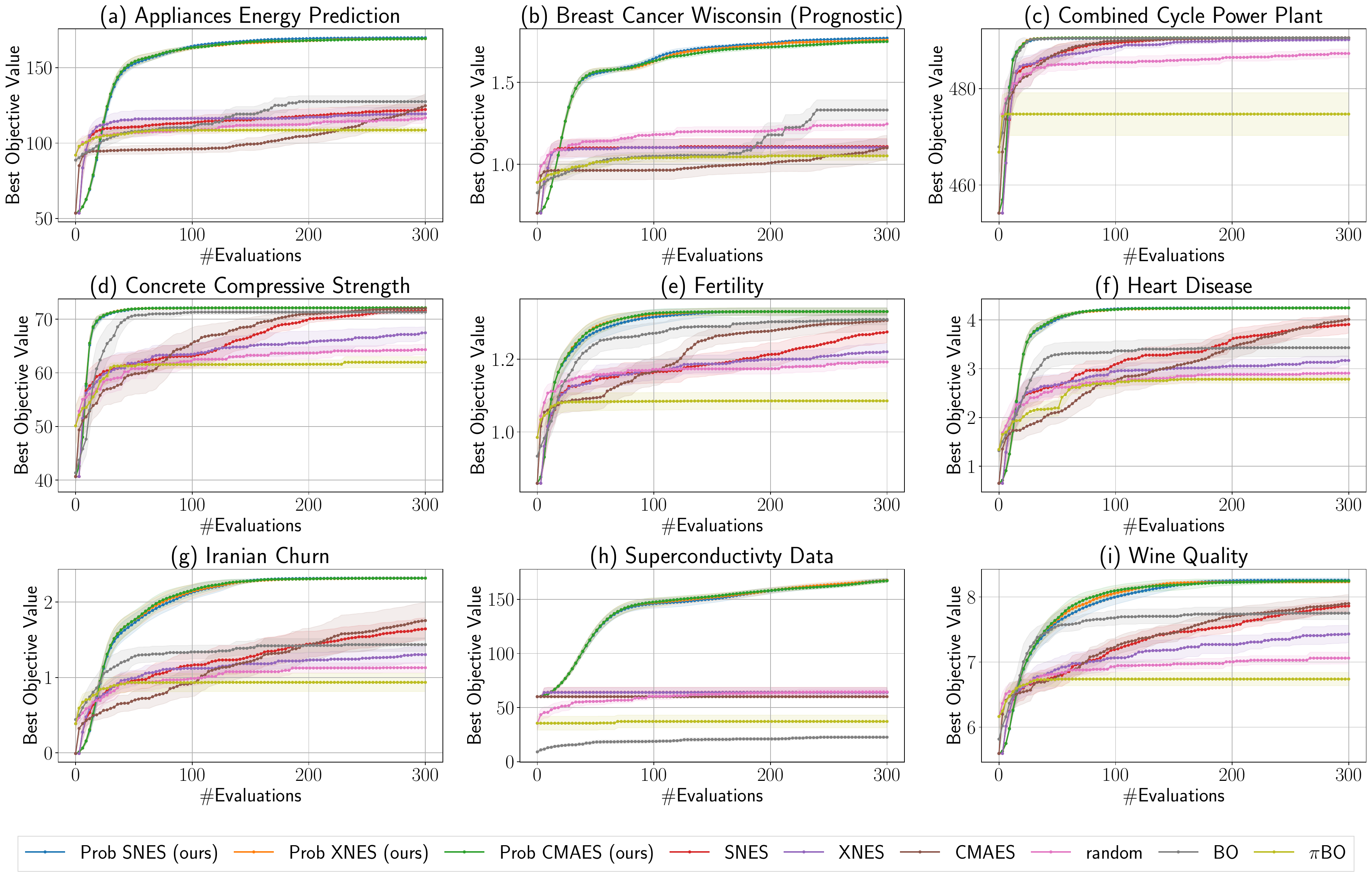}
    \vspace{-2em}
    \caption{UCI semisupervised optimisation}
    \label{fig:uci}
\end{figure*}
\textbf{Baseline.}
Our goal is to showcase the performance improvement of ProbNES compared to their non-probabilistic counterpart. In particular, we derived ProbNES algorithms for {\fontfamily{cmtt}\selectfont CMA-ES} \citep{hansen2016cma}, {\fontfamily{cmtt}\selectfont XNES} and {\fontfamily{cmtt}\selectfont SNES} \cite{wierstra2014natural}. As a result, we compare against these classical evolutionary strategy algorithm. Furthermore, we also added two baselines, {\fontfamily{cmtt}\selectfont $\pi$BO} \citep{hvarfner2022pi} and vanilla {\fontfamily{cmtt}\selectfont BO} \citep{garnett2023bayesian}. We added these last two baselines to showcase how the more exploitative nature of ProbNES can lead to better performance than sample efficient global method, even when they are augmented by prior knowledge.

\textbf{Tasks.} 
We conduct our evaluation on synthetic and real-world data. First we use synthetic test functions. We proceed to evaluate on UCI datasets \citep{markuci} used for regression tasks and transformed into optimisation tasks. Then, we benchmark on latent space optimisation by finding particular classes in MNIST \citep{deng2012mnist} and CIFAR10 \citep{cifar10} via optimising the latent space of pre-trained generators. We also perform hyperparameter tuning with prior belief based on default parameter values on fcnet from the Profet suite \citep{klein2019meta} and of a GP through its marginal log-likelihood. Finally, we learn linear policy from locomotion tasks.

\textbf{Metric.} 
We adopt the simple regret as the performance metric:
\begin{align}
    R_T := \max_{t \in [T]} f(x^*) - f(x_t), 
\end{align}
where $[T] := [1,\cdots,T]$. When $R_T = 0$, it follows that $x^* = x_T$. Thus, this regret metric is equivalent to our objective in Problem~(\ref{eq:prob_to_solve}). 

In our plots, we show the regret of the objective values as a function of the number of queries with error bars indicating $95 \%$ confidence intervals and based on $15$ experiments. We used an $16$ cores 3.80GHz Intel CPU with 64 GB of RAM available as well as a NVIDIA GeForce RTX 4070 SUPER GPU with 12 GB memory. Our implementation is available at \url{https://anonymous.4open.science/r/semisupervised-5A83} where all experiments can be reproduced.

\subsection{Test function optimisation} \label{subsection:exp_testfunctions}
\begin{figure*}[h]
    \centering
\includegraphics[width=\textwidth]{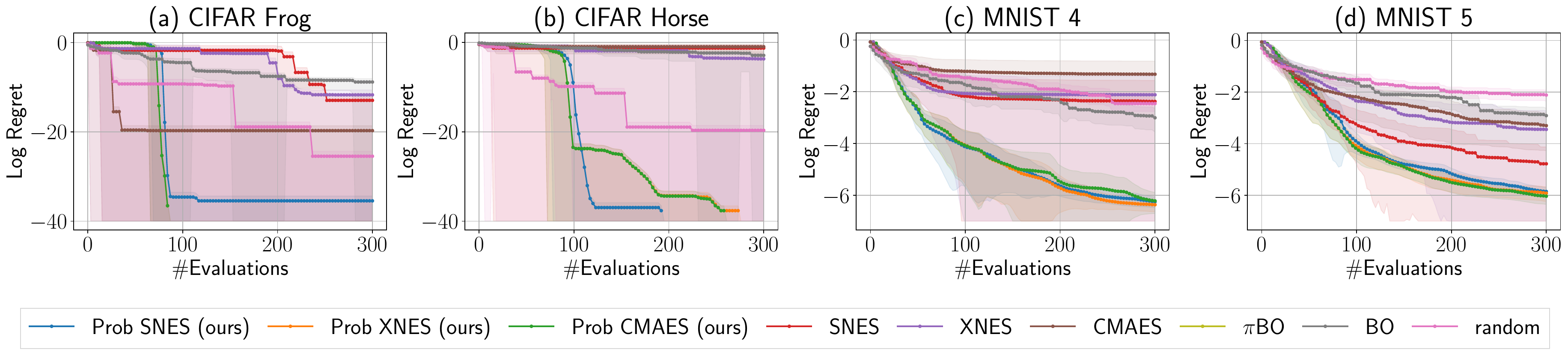}
    \vspace{-2em}
    \caption{Latent Space Optimization}
    \label{fig:lso}
\end{figure*}
 \begin{figure*}[h]
    \centering
\includegraphics[width=\textwidth]{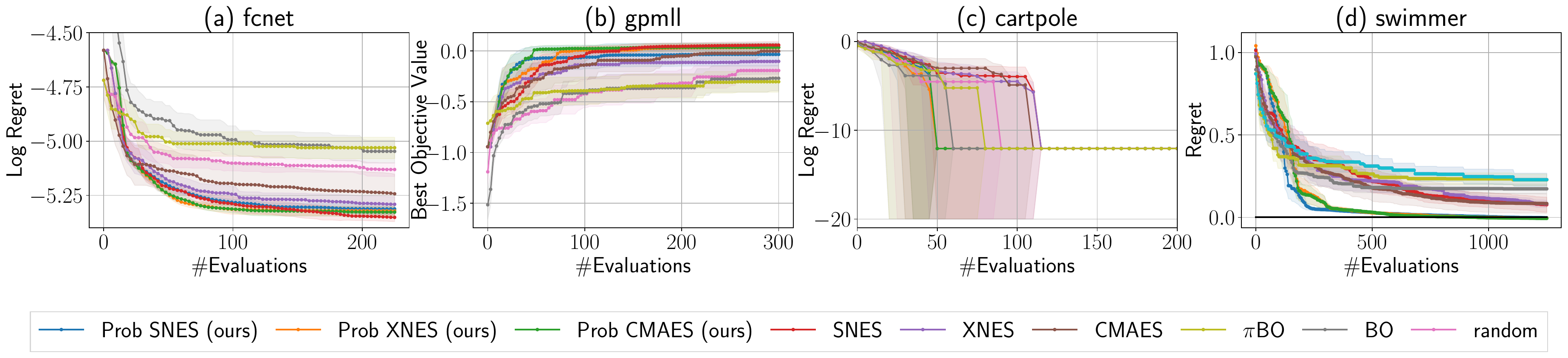}
    \vspace{-2em}
    \caption{Hyperparameter tuning and locomotion task}
    \label{fig:hp}
\end{figure*}
In this experiment, we selected eight classical test functions: Ackley, Rastrigin, Branin, Griewank, Levy, Shekel, StyblinskyTang and ThreeHumpCamel \citep{simulationlib}. Details about their dimensionality, analytical forms and optimum are given in \ref{appendix:test_function}. We chose an initial distribution $\nu_{\theta_0} = \mathcal{N}\left(\boldsymbol{-1}, \boldsymbol{I}\right)$ which is not centered on the test function optimum but are within two standard deviations. For {\fontfamily{cmtt}\selectfont $\pi$BO} and {\fontfamily{cmtt}\selectfont BO}, we also need an input bounding box, which we choose to be $\left[ -3, 3\right]^{d}$ as it corresponds to three standard deviation of our initial distribution.

Results are given in Figure \ref{fig:test_function}, where our method consistently outperforms the other baselines. We can see {\fontfamily{cmtt}\selectfont BO} usually outperforms traditional NES algorithms and {\fontfamily{cmtt}\selectfont $\pi$BO}, except for (f) where they give similar results.

\subsection{Semi-supervised Task}
In this task, we took nine regression datasets, namely Appliances Energy Prediction, Breast Cancer Wisconsin (Prognostic), Combined Cycle Power Plant, Concrete Compressive Strength, Fertility, Heart Disease, Iranian Churn, Superconductivity Data and Wine Quality. Detailed descriptions are available in \ref{appendix:uci}. A Support Vector Machine model with RBF kernel is fitted onto the data as well as a multivariate Gaussian distribution on the input data. We then transform each task into an optimisation task via optimising this surrogate model with the fitted normal distribution as the initial distribution.

Results are shown in Figure \ref{fig:uci} with the best observed value $\max_{t \in [T]} f(x_t)$. In this example, our method successfully exploit this initial data distribution to find an optimal point faster than its counterparts.
\label{subsection:exp_semisupervised}

\subsection{Latent space optimisation Task} \label{subsection:exp_latentspace}
In this task, we selected two labels in MNIST and CIFAR10 datasets. Furthermore, for each dataset, we selected a pretrained generative model (VAE for MNIST, GAN for CIFAR10\footnote{The weights and architectures are available \href{https://github.com/csinva/gan-vae-pretrained-pytorch/tree/master}{here}.}) and classifiers. We aimed at optimising a variable in the latent space that produced samples with maximum probability to belong to the selected classes according to the pretrained classifier. Details about the objective function are given in \ref{appendix:lso}.

Results are shown in Fig.~\ref{fig:lso} with regret in log scale. As we can see from the performance of the random search, the task seems not complex enough in the case of CIFAR10, which can be the result of overestimating class probabilities in the pretrained classifier. In any cases, our method outperform the baseline algorithms.

\subsection{Hyperparameter tuning with prior information \& locomotion tasks}
\label{subsection:exp_hyperparameter}
We conducted real-world hyperparameter tuning experiment from the Profet suite \citep{klein2019meta}. In particular, we took the FC-Net(6D) and SVM(2D) tasks. In addition, we also formulated a hyperparameter tuning experience whereby the hyperparameters of a GP, namely likelihood noise, mean constant, outputscale and lengthscale, are optimised via the marginal log-likelihood. Results are shown in Fig.~\ref{fig:hp} (a) and (b), and experimental details in \ref{appendix:hpo}.

We also carried out experiments from MuJoCo locomotion tasks \citep{todorov2012mujoco} a linear policy mapping states to actions is optimised to maximise the reward received from the environment. We use the environments CartPole-v1 (4D) as well as Swimmer-v4 (16D). The results are shown in Fig.~\ref{fig:hp} (c) and (d) and
the experimental details are described in \ref{appendix:rl}.

\section{Conclusion and Future Work}

In this paper, we improved the sample efficiency of NES algorithms, a subclass of zeroth order optimisation methods that refine a search distribution to attain the optimum of a given objective function. We considered the IGO perspective of NES algorithms, whereby parameter updates of the search distribution are cast as a natural gradient descent step with respect to the Fisher metric in the space of probability distributions.

Through this IGO perspective, we were able to exploit BQ to formulate probabilistic versions of NES algorithms, where samples can be actively gathered and where the natural gradient is estimated via BQ. 

We showed through synthetic test function optimisation, data informed optimisation of objectives from the UCI datasets, latent space optimisation, hyperparameter optimisation and locomotion tasks that our method outperforms its non-probabilistic counterparts.

We focused on a single multivariate normal distribution as the search distribution for closed form updates. However, if we only estimate the Fisher matrix, it is possible to exploit more complex distributions such as mixture of multivariate Gaussians or deep generative models such as VAEs, GANs, or diffusion models \citep{kahn2023bayesian}.

\bibliography{ressources}

\appendix
\onecolumn

\section{Gaussian Formulae}

In this section we provide formulae regarding the multivariate normal distribution that are required in the computation of the covariance structure in the probabilistic method of lines.
\subsection{Multivariate Normal Distribution Density}
The probability density function of a multivariate normal distribution $\Xb \sim \mathcal{N}(\mub, \Sigmab)$ in $\mathbb{R}^{D}$ is:
\begin{equation}
    \nu_{\mub, \Sigmab}(\x) = \frac{1}{\sqrt{|2\pi\Sigmab|}} e^{ -\frac{1}{2}(\x - \mub)^{T}\Sigmab^{-1}(\x - \mub)}
\end{equation}
\subsection{Multivariate Normal Distribution Density under Linear Transformations}
Let $\Xb \sim \mathcal{N}(\mub, \Sigmab)$, then, the linear transformation of $\Xb$ is also normally distributed: $$\mathbf{Y} = \mathbf{A}\Xb + \mathbf{b} \sim \mathcal{N}(\mathbf{A}\mub + \mathbf{b}, \mathbf{A}\Sigmab\mathbf{A}^{T})$$

\begin{proof}
    See section 8.1.4 in \cite{petersen2008matrix}.
\end{proof}
\subsection{Product of Two Normal Densities}
Let $\nu_1(\x) = \mathcal{N}(\x; \mub_1, \Sigmab_1)$ and $\nu_2(\x) = \mathcal{N}(\x; \mub_2, \Sigmab_2)$, we have:
\begin{equation}
    \nu_1(\x) \nu_2(\x) = \frac{1}{\left(2\pi\right)^{D} \sqrt{|\Sigmab_1||\Sigmab_2|}} \times \exp \left(-\frac{1}{2} \left( (\x - \nub)^{T} \Pib^{-1} (\x - \nub) - \nub^{T}\Pib^{-1}\nub + \mub_1^{T}\Sigmab_{1}^{-1} \mub_1 + \mub_2^{T}\Sigmab_{2}^{-1} \mub_2 \right) \right)
\end{equation}
With
\begin{equation*}
\begin{split}
    \Pib &= \left( \Sigmab_{1}^{-1} + \Sigmab_{2}^{-1}\right)^{-1} = \Sigmab_{1}\left( \Sigmab_{1} + \Sigmab_{2}\right)^{-1}\Sigmab_{2} = \Sigmab_{2}\left( \Sigmab_{1} + \Sigmab_{2}\right)^{-1}\Sigmab_{1} \\
    \nub &= \Pib \Sigmab_{1}^{-1}\mub_1 + \Pib \Sigmab_{2}^{-1}\mub_2 \\
    &= \Sigmab_{2}\left(\Sigmab_{1} + \Sigmab_{2}\right)^{-1}\mub_{1} + \Sigmab_{1}\left(\Sigmab_{1} + \Sigmab_{2}\right)^{-1}\mub_{2}
\end{split}
\end{equation*}
This formulation can be rewritten as:
\begin{equation} \label{eq:productgauss}
\begin{split}
    \nu_1(\x) \nu_2(\x) &= \frac{1}{\left(2\pi\right)^{D} \sqrt{|\Sigmab_1||\Sigmab_2|}} \times \exp \left(-\frac{1}{2} \left( \left(\x - \nub\right)^{T} \Pib^{-1} \left(\x - \nub\right) + \left(\mub_1 - \mub_2\right)^{T} \left(\Sigmab_1 + \Sigmab_2\right)^{-1} \left(\mub_1 - \mub_2\right) \right)\right) \\
    &= \mathcal{N}(\x; \nub,    \Pib) \mathcal{N}(\mub_1; \mub_2, \Sigmab_{1} + \Sigmab_{2})
\end{split}
\end{equation}
\begin{proof}
    See section 8.1.8 from \cite{petersen2008matrix}.
\end{proof}
\subsection{Integration of Two Multivariate Normal Distributions}
Let $\nu_1(\x) = \mathcal{N}(\x; \mub_1, \Sigmab_1)$ and $\nu_2(\x) = \mathcal{N}(\x; \mub_2, \Sigmab_2)$, we have:
\begin{equation} \label{eq:productint}
    \int_{\mathbb{R}^{D}} \nu_1(\x) \nu_2(\x)d\x = \mathcal{N}\left(\mub_1;\mub_2, \Sigmab_1 + \Sigmab_2\right)
\end{equation}
\begin{proof}
    Immediate result after integrating the first term in equation \ref{eq:productgauss}.
\end{proof}

\subsection{First Moment of Products of Multivariate Normal Distributions}
Let $\nu_1(\x) = \mathcal{N}(\x; \mub_1, \Sigmab_1)$ and $\nu_2(\x) = \mathcal{N}(\x; \mub_2, \Sigmab_2)$, we have:
\begin{equation}
\begin{split}
    \int_{\mathbb{R}^{D}} \x \nu_1(\x) \nu_2(\x)d\x &= \nub \mathcal{N}\left(\mub_1;\mub_2, \Sigmab_1 + \Sigmab_2\right) \\
    &= \Pib \Sigmab_{1}^{-1}\mub_1 \mathcal{N}\left(\mub_1;\mub_2, \Sigmab_1 + \Sigmab_2\right) + \Pib \Sigmab_{2}^{-1}\mub_2 \mathcal{N}\left(\mub_2;\mub_1, \Sigmab_1 + \Sigmab_2\right)
\end{split}
\end{equation}
With $\nub$ defined in A.2.
\begin{proof}
\begin{equation}
\begin{split}
    \int_{\mathbb{R}^{D}} \x \nu_1(\x) \nu_2(\x)d\x &= \left( \underbrace{\int_{\mathbb{R}^{D}} \x \mathcal{N}(\x; \nub, \Pib) d\x}_{=\mathbb{E}[X], X \sim \mathcal{N}(\nub, \Pib)} \right) \mathcal{N}(\mub_1; \mub_2, \Sigmab_{1} + \Sigmab_{2}) \\
    &= \nub \mathcal{N}(\mub_1; \mub_2, \Sigmab_{1} + \Sigmab_{2})
\end{split}
\end{equation}
\end{proof}
\subsection{Second Moment of Products of Multivariate Normal Distributions}
Let $\nu_1(\x) = \mathcal{N}(\x; \mub_1, \Sigmab_1)$, $\nu_2(\x) = \mathcal{N}(\x; \mub_2, \Sigmab_2)$, and $\otimes$ the outer product operator. We have:
\begin{equation}
    \int_{\mathbb{R}^{D}} \left( \x \otimes \x \right) \nu_1(\x) \nu_2(\x)d\x = \left( \Pib + \nub \otimes \nub \right)\mathcal{N}\left(\mub_1;\mub_2, \Sigmab_1 + \Sigmab_2\right)
\end{equation}
With 
\begin{equation*}
    \Pib + \nub \nub^{T} = \Pib + \Pib \Sigmab_{1}^{-1}\mub_1 \mub_{1}^{T} \Sigmab_{1}^{-1} \Pib + \Pib \Sigmab_{2}^{-1}\mub_2 \mub_{2}^{T} \Sigmab_{2}^{-1} \Pib + \Pib \Sigmab_{1}^{-1}\mub_1 \mub_{2}^{T} \Sigmab_{2}^{-1} + \Pib \Sigmab_{2}^{-1}\mub_2 \mub_{1}^{T} \Sigmab_{1}^{-1} \Pib
\end{equation*}
\begin{proof}
\begin{equation}
\begin{split}
     & \int_{\mathbb{R}^{D}} \left( \x \otimes \x \right) \nu_1(\x) \nu_2(\x)d\x = \left( \underbrace{\int_{\mathbb{R}^{D}} \left( \x \otimes \x \right) \mathcal{N}(\x; \nub, \Pib) d\x}_{=(1)} \right) \mathcal{N}(\mub_1; \mub_2, \Sigmab_{1} + \Sigmab_{2}) \\
    (1) &= \int_{\mathbb{R}^{D}} \left( \left(\x - \nub \right) \otimes \left(\x - \nub \right) + \left(\x - \nub \right) \otimes \nub + \nub \otimes \left(\x - \nub \right) + \nub \otimes \nub\right) \mathcal{N}(\x; \nub, \Pib) d\x \\
    &= \Pib + \nub \otimes \nub
\end{split}
\end{equation}
\end{proof}
\subsection{Third Moment of Products of Multivariate Normal Distributions} \label{appendix:thirdmoment}
Let $\nu_1(\x) = \mathcal{N}(\x; \mub_1, \Sigmab_1)$, $\nu_2(\x) = \mathcal{N}(\x; \mub_2, \Sigmab_2)$, and $\otimes$ the outer product operator. We have:
\begin{equation}
    \int_{\mathbb{R}^{D}} \left( \x \otimes \x \otimes \x \right) \nu_1(\x) \nu_2(\x)d\x = \left(\nub \otimes \nub \otimes \nub + \sum\limits_{\tau \in \Taub_3} \tau(\nub \otimes \Pib)\right) \mathcal{N}\left(\mub_1;\mub_2, \Sigmab_1 + \Sigmab_2\right)
\end{equation}
Where $\nub$ and $\Pib$ are defined in A.2 and $\Taub_3 = \left\{ \tau_c : c \in \mathcal{C}_{3} \right\}$, where $\mathcal{C}_{3}$ is the set of cyclic permutation over $[3]$, is the set of operators over 3D tensor that perform cyclic permutation of the indices.
\begin{proof}
\begin{equation}
\begin{split}
    \x \otimes \x \otimes \x &= \left(\x - \nub \right) \otimes \left(\x - \nub \right) \otimes \left(\x - \nub \right) \\
    &+ \nub \otimes \left(\x - \nub \right) \otimes \left(\x - \nub \right) + \left(\x - \nub \right) \otimes \nub \otimes \left(\x - \nub \right) + \left(\x - \nub \right) \otimes \left(\x - \nub \right) \otimes \nub \\
    &+ \nub \otimes \nub \otimes \left(\x - \nub \right) + \nub \otimes \left(\x - \nub \right) \otimes \nub + \left(\x - \nub \right) \otimes \nub \otimes \nub \\
    &+ \nub \otimes \nub \otimes \nub \\
\end{split}
\end{equation}
The first and third line integrates to zero as every term of the 3D tensor involves an odd number of the dimensions of the centered multivariate random variable $\Xb - \mub$, which gives the result.
\end{proof}
\subsection{Derivative of multivariate normal distribution with respect to its parameters}
Let $\nu(\x) = \sum\limits_{i \in [K]} \alpha_i \mathcal{N}(\mub_i, \Sigmab_i)$:
\begin{equation}
\begin{split}
    \frac{\partial \ln{\nu(\x)} }{\partial \alpha_j} &= \frac{\alpha_j \mathcal{N}(\mub_j, \Sigmab_j)}{\sum\limits_{i \in [K]} \alpha_i \mathcal{N}(\mub_i, \Sigmab_i)} \frac{1}{\alpha_j} \\
    \frac{\partial \ln{\nu(\x)} }{\partial \mub_j} &= \frac{\alpha_j \mathcal{N}(\mub_j, \Sigmab_j)}{\sum\limits_{i \in [K]} \alpha_i \mathcal{N}(\mub_i, \Sigmab_i)} \big[\Sigmab_j^{-1}(\x - \mub_j) \big] \\
    \frac{\partial \ln{\nu(\x)} }{\partial \Sigmab_j} &= \frac{\alpha_j \mathcal{N}(\mub_j, \Sigmab_j)}{\sum\limits_{i \in [K]} \alpha_i \mathcal{N}(\mub_i, \Sigmab_i)} \frac{1}{2} \big[-\Sigmab_j^{-1} +\Sigmab_j^{-1}(\x - \mub_j)(\x - \mub_j)^{T}\Sigmab_j^{-1} \big]
\end{split}
\end{equation}
\begin{proof}
    See section 8.4.2 in \cite{petersen2008matrix}.
\end{proof}

\section{Bayesian Quadrature formula} \label{appendix:formula}
\subsection{Quadrature}
\textbf{Multivariate Normal Distribution.} \\
If $\nu(\x) = \mathcal{N}(\x; \mub, \Sigmab)$ and $k(\x, \x') = \theta \mathcal{N}(\x; \x', \Lambda)$ and $f \sim \mathcal{GP}(\mb, \kb)$ then closed form solution are available for the computation of the distribution of F:
\begin{equation}
\begin{split}
    \tb(\x_{i}) & = \theta \mathcal{N}(\x_{i}; \mub,\Sigmab + \Lambda) \\
    \mathbf{R} & = \theta (\det 2\pi(2\Sigmab + \Lambda))^{-\frac{1}{2}}
\end{split}
\end{equation}
\textbf{Mixture of Multivariate Normal Distribution.} \\
If $\nu(\x) = \sum\limits_{i \in [K]} \alpha_{i} \mathcal{N}(\x; \mub_i, \Sigmab_i)$ and $k(\x, \x') = \mathcal{N}(\x; \x', \Lambda)$ and $f \sim \mathcal{GP}(0, \kb)$ then:
\begin{equation}
\begin{split}
    \tb(\x_{i}) & = \sum\limits_{i \in [K]} \alpha_i \mathcal{N}(\x_{i}; \mub_i, \Sigmab_i + \Lambda) \\
    \mathbf{R} & = \sum\limits_{i, j \in [K]} \alpha_i \alpha_j(\det 2\pi(\Sigmab_i + \Sigmab_j + \Lambda))^{-\frac{1}{2}}
\end{split}
\end{equation}

\subsection{Covariance Structure of Quadrature Gradient}
\label{appendix:probaline}

We recall the covariance structure of the gradient vector of the integral of the objective function with respect to the search distribution in the space of distribution parameters from \ref{eq:covariancestructure}
\begin{equation}
\begin{split}
    \mathbb{E}(g(\theta)) &= \int_{\X} \mathbf{m}_{post}(\x)d\nu_{\theta}(\x)\\
    &= \m + \mathbf{t}_{kX}^{T}\left[\kb_{XX} + \sigma_{noise}^{2}\mathbf{I}\right]^{-1} \left(\mathbf{Y} - \mathbf{m}(\Xb) \right)\\
    \mathbb{E}(\nabla_{\theta} g(\theta)) &= \int_{\X} \nabla_{\theta} \ln{\nu_{\theta}}|_{\theta}(\x) \mathbf{m}_{post}(\x)d\nu_{\theta}(\x)\\
    &= \boldsymbol{\mathcal{M}} + \boldsymbol{\mathcal{T}}_{kX}^{T}\left[\kb_{XX} + \sigma_{noise}^{2}\mathbf{I}\right]^{-1} \left(\mathbf{Y} - \mathbf{m}(\Xb) \right)\\
    \text{Cov}(g(\theta_1), g(\theta_2)) &= \iint_{\X} \kb_{post}(\x, \x')d\nu_{\theta_1}(\x)d\nu_{\theta_2}(\x') \\
    &= \mathbf{R}_{12} - \mathbf{t}_{1X}^{T}\left[\kb_{XX} + \sigma_{noise}^{2}\mathbf{I}\right]^{-1}\mathbf{t}_{2X}\\
    \text{Cov}(g(\theta_1), \nabla_{\theta} g(\theta_2)) &= \iint_{\X} \nabla_{\theta} \ln{\nu_{\theta}}|_{\theta_{2}}(\x') \kb_{post}(\x, \x') d\nu_{\theta_1}(\x)d\nu_{\theta_2}(\x') \\
    &= \mathbf{R}_{12'} - \mathbf{t}_{1X}^{T}\left[\kb_{XX} + \sigma_{noise}^{2}\mathbf{I}\right]^{-1} \boldsymbol{\mathcal{T}}_{2X}\\
    \text{Cov}(\nabla_{\theta} g(\theta_1),  \nabla_{\theta} g(\theta_2)) &= \iint_{\X} \nabla_{\theta} \ln{\nu_{\theta}}|_{\theta_1}(\x) \otimes \nabla_{\theta} \ln{\nu_{\theta}}|_{\theta_2}(\x') \kb_{post}(\x, \x') d\nu_{\theta_1}(\x)d\nu_{\theta_2}(\x') \\
    &= \mathbf{R}_{1'2'} - \boldsymbol{\mathcal{T}}_{1X}^{T}\left[\kb_{XX} + \sigma_{noise}^{2}\mathbf{I}\right]^{-1}\boldsymbol{\mathcal{T}}_{2X}
\end{split}
\end{equation}
We define $\nu_{\theta_1}(\x) = \mathcal{N}(\x; \mub_1, \Sigmab_1)$, $\nu_{\theta_t}(\x) = \mathcal{N}(\x; \mub_2, \Sigmab_2)$, $k(\x, \x') = \mathcal{N}(\x; \x', \Lambda)$, $\m(\x) = \mathbf{v} \in \mathbb{R}^{d}$ and $f \sim \mathcal{GP}(\m, \kb)$. We suppose we have a data set $\mathcal{D}$ of points observed from the evaluation function. Let $\Gammab = \Lambda + \Sigmab_1 + \Sigmab_2$ and $\gammab = \Gammab^{-1}(\mub_1 - \mub_2)$, $\Taub_3$ defined in \ref{appendix:thirdmoment}, $\Taub = \left\{\left(1, 2, 3, 4 \right), \left(2, 1, 3, 4 \right), \left(1, 2, 4, 3 \right), \left(2, 1, 4, 3 \right), \left(2, 3, 1, 4 \right), \left(1, 4, 2, 3 \right) \right\}$, and $\boldsymbol{\mathcal{S}} = \left\{\left(1, 2, 3, 4 \right), \left(1, 3, 2, 4 \right), \left(1, 4, 2, 3 \right) \right\}$. Then, the quantities involved in eq \ref{eq:covariancestructure} are:
\begin{subequations}
\begin{align}
    \mb &= \mathbf{v}  \label{eq:Ea}\\
    \boldsymbol{\mathcal{M}} &= \boldsymbol{0} \label{eq:Eb}\\
    \mathbf{R}_{12} &= \mathcal{N}(\mub_{1}; \mub_2, \Gammab)\label{eq:Ec}\\
    \mathbf{R}_{12'\mub} &= \gammab \mathcal{N}(\mub_1; \mub_2, \Gammab)\label{eq:Ed}\\
    \mathbf{R}_{12'\Sigmab} &= -\frac{1}{2}\left(\Gammab^{-1} - \gammab \otimes \gammab\right) \mathcal{N}(\mub_1; \mub_2, \Gammab)\label{eq:Ee}\\
    \mathbf{R}_{1'2'\mub\mub} &=\left(\Gammab^{-1} - \gammab \otimes \gammab\right) \mathcal{N}(\mub_1; \mub_2, \Gammab)\label{eq:Ef}\\
    \mathbf{R}_{1'2'\mub\Sigmab} &= \frac{1}{2} \left( \sum\limits_{\tau \in \Taub_3} \tau(\gammab \otimes \Gammab^{-1}) - \gammab \otimes \gammab \otimes \gammab \right)\mathcal{N}(\mub_1; \mub_2, \Gammab)\label{eq:Eg}\\
    \mathbf{R}_{1'2'\Sigmab\Sigmab} &= \frac{1}{4}\left(\gammab \otimes \gammab \otimes \gammab \otimes \gammab - \sum\limits_{\tau \in \Taub} \left[ \gammab \otimes \Gammab^{-1} \otimes \gammab \right]_{\tau} + \sum\limits_{\sigma \in \boldsymbol{\mathcal{S}}} \left[ \Gammab^{-1}\otimes \Gammab^{-1} \right]_{\sigma} \right) \mathcal{N}(\mub_1; \mub_2, \Gammab)\label{eq:Eh}\\
    \mathbf{t}_{kX} &= \mathcal{N}(\Xb; \mub_k, \Lambda + \Sigmab_k)\label{eq:Ei}\\
    \left[\boldsymbol{\mathcal{T}}_{kX}\right]_{i, \mub} &= \left(\Lambda + \Sigmab_k\right)^{-1} \left(\x_i - \mub_k\right) \mathcal{N}(\x_i; \mub_k, \Lambda + \Sigmab_k)\label{eq:Ej}\\
    \left[\boldsymbol{\mathcal{T}}_{kX}\right]_{i, \Sigmab} &= \frac{1}{2} \left(\Lambda + \Sigmab_k\right)^{-1}\left(\left(\x_i - \mub_k \right)\left(\x_i - \mub_k \right)^{T} - \left(\Lambda + \Sigmab_k\right)\right)\left(\Lambda + \Sigmab_k\right)^{-1} \mathcal{N}(\x_i; \mub_k, \Lambda + \Sigmab_k) \label{eq:Ek}
\end{align}
\end{subequations}

\begin{proof} Let $D \in \mathbb{N}^{*}$, $D \geq k \in \mathbb{N}^{*}$, $\left( d_i \right)_{1\leq i \leq D} \in \mathbb{N}^{*}$. Let's define the matrix product over tensor dimensions:
\begin{equation}
\begin{split}
    *_{k} : \mathbb{R}^{d_k \times d_k} \times \mathbb{R}^{\prod\limits_{i = 1}^{D} d_i} & \rightarrow \mathbb{R}^{\prod\limits_{i = 1}^{D} d_i} \\
     \left( \Sigmab, \mathcal{T} \right) &\mapsto  \Tilde{\mathcal{T}}\\
\end{split}
\end{equation}
with
\begin{equation}
    \left[\Tilde{\mathcal{T}} \right]_{i_1, \dots, i_D} = \sum\limits_{l = 1}^{d_k} \Sigmab_{i_k, l} \left[\mathcal{T} \right]_{i_1, \dots, i_{k-1}, l, \dots, i_{k+1},\dots,  i_D}
\end{equation}
Let's now derive all the equations:\\
\ref{eq:Ea}:
\begin{equation*}
\m_t = \int_{\X} \m(\x) d\nu_{\theta_{t}}(\x) = \m \int_{\X} d\nu_{\theta_{t}}(\x) = \m
\end{equation*}
\ref{eq:Eb}: 
\begin{equation*}
\boldsymbol{\mathcal{M}} = \int_{\X} \nabla_{\theta} \ln{\nu_{\theta}}|_{\theta}(\x) \m(\x) d\nu_{\theta}(\x) = \m \int_{\X} \nabla_{\theta} \nu_{\theta}|_{\theta}(\x) d\x = \m \nabla_{\theta} 1 = 0
\end{equation*}
\ref{eq:Ec}:
\begin{equation*}
\begin{split}
\mathbf{R}_{12} &= \iint_{\X} \kb(\x, \x')d\nu_{\theta_{t_1}}(\x)d\nu_{\theta_{t_2}}(\x') = \iint_{\X} \mathcal{N}(\x; \x', \Lambda) \mathcal{N}(\x; \mub_1, \Sigmab_1) \mathcal{N}(\x'; \mub_2, \Sigmab_2)d\x d\x' \\ 
&= \int_{\X}  \mathcal{N}(\x'; \mub_1, \Lambda + \Sigmab_1) \mathcal{N}(\x'; \mub_2, \Sigmab_2)d \x' = \mathcal{N}(\mub_1; \mub_2, \Gammab)
\end{split}
\end{equation*}
\ref{eq:Ed}:
\begin{equation*}
\begin{split}
\mathbf{R}_{12'\mub} &= \iint_{\X} \nabla_{\theta} \ln{\nu_{\theta}}|_{\theta_{t_2}}(\x') \kb(\x, \x') d\nu_{\theta_{t_1}}(\x)d\nu_{\theta_{t_2}}(\x') = \iint_{\X} \big[\Sigmab_2^{-1}(\x - \mub_2) \big] \kb(\x, \x') d\nu_{\theta_{t_1}}(\x)d\nu_{\theta_{t_2}}(\x') \\
&= \Sigmab_2^{-1} \int_{\X} \mathcal{N}(\x; \mub_1, \Sigmab_1) \int_{\X} \x' \mathcal{N}(\x'; \x - \mub_2, \Lambda) \mathcal{N}(\x'; \mathbf{0}, \Sigmab_2) d \x' d \x \\
&= \Sigmab_2^{-1} \int_{\X} \Sigmab_2 \left( \Sigmab_2 + \Lambda \right)^{-1} \left(\x - \mub_{2}\right) \mathcal{N}(\x - \mub_2; \mathbf{0}, \Sigmab_2 + \Lambda) \mathcal{N}(\x; \mub_1, \Sigmab_1) d \x \\
&= \left(\Sigmab_1 + \Sigmab_2 + \Lambda \right)^{-1} \left( \mub_1 - \mub_2 \right) \mathcal{N}(\mub_1; \mub_2,  \Sigmab_1 + \Sigmab_2 + \Lambda) \\
\end{split}
\end{equation*}
\ref{eq:Ee}:
\begin{equation*}
\begin{split}
\mathbf{R}_{12'\Sigmab} &= \iint_{\X} \nabla_{\Sigmab} \ln{\nu_{\theta}}|_{\theta_{t_2}}(\x') \kb(\x, \x') d\nu_{\theta_{t_1}}(\x)d\nu_{\theta_{t_2}}(\x') \\
&= \iint_{\X} \frac{1}{2}\big[-\Sigmab_2^{-1} +\Sigmab_2^{-1}(\x - \mub_2)(\x - \mub_2)^{T}\Sigmab_2^{-1} \big] \kb(\x, \x') d\nu_{\theta_{t_1}}(\x)d\nu_{\theta_{t_2}}(\x') \\
&= -\frac{1}{2}\Sigmab_2^{-1} \mathbf{R}_{12} + \frac{1}{2} \Sigmab_2^{-1} \underbrace{\iint_{\X} (\x - \mub_2)(\x - \mub_2)^{T} \kb(\x, \x') d\nu_{\theta_{t_1}}(\x)d\nu_{\theta_{t_2}}(\x')}_{(1)} \Sigmab_2^{-1} \\
(1) &= \int_{\X} \left( \Pib_{2} + \nub_{2} \otimes \nub_{2} \right) \mathcal{N}(\x; \mub_2, \Sigmab_2 + \Lambda) \mathcal{N}(\x; \mub_1,  \Sigmab_1) d \x \text{ with } \left\{
    \begin{array}{ll}
        \Pib_{2} = \left( \Lambda^{-1} + \Sigmab_{2}^{-1} \right)^{-1} \\
        \nub_{2} = \Sigmab_{2}\left( \Lambda + \Sigmab_{2} \right)^{-1}\left( \x - \mub_{2}\right)
    \end{array}
\right. \\
&= \Pib_2 \mathcal{N}(\mub_1; \mub_2, \Sigmab_1 + \Sigmab_2 + \Lambda) \\
&+ \Sigmab_{2}\left( \Lambda + \Sigmab_{2} \right)^{-1} \underbrace{\int_{\X} (\x - \mub_2)(\x - \mub_2)^{T} \mathcal{N}(\x; \mub_1,  \Sigmab_1) \mathcal{N}(\x; \mub_2,  \Lambda + \Sigmab_2) d \x}_{(2)} \left( \Lambda + \Sigmab_{2} \right)^{-1} \Sigmab_{2} \\
(2) &= \int_{\X} \x \x^{T} \mathcal{N}(\x; \mub_1 - \mub_2,  \Sigmab_1) \mathcal{N}(\x; \Lambda + \Sigmab_2) d \x \\
&= \left( \Pib_{1} + \nub_{1} \otimes \nub_{1} \right) \mathcal{N}(\mub_1; \mub_2, \Sigmab_1 + \Sigmab_2 + \Lambda) \text{ with } \left\{
    \begin{array}{ll}
        \Pib_{1} = \Sigmab_{1} \left(\Sigmab_1 + \Sigmab_2 + \Lambda \right)^{-1}\left( \Sigmab_2 + \Lambda \right) \\
        \nub_{1} = \Pib_{1} \Sigmab_{1}^{-1}\left( \mub_{1} - \mub_{2}\right)
    \end{array}
\right. \\
\mathbf{R}_{12'\Sigmab} &= \frac{1}{2}\left[ \underbrace{-\Sigmab_{2}^{-1} + \Sigmab_{2}^{-1} \left( \Pib_2 + \Pib_2 \Lambda^{-1} \left( \Pib_1 + \nub_1 \otimes \nub_1 \right) \Lambda^{-1} \Pib_2 \right) \Sigmab_{2}^{-1}}_{(3)} \right] \mathcal{N}(\mub_1; \mub_2, \Gammab)\\
(3) &= \underbrace{-\Sigmab_{2}^{-1} + \left( \Sigmab_{2} + \Lambda \right)^{-1} \Lambda \Sigmab_{2}^{-1}}_{= -\left( \Sigmab_{2} + \Lambda \right)^{-1}} + \left( \Sigmab_{2} + \Lambda \right)^{-1} \Sigmab_1 \Gammab^{-1} + \Gammab^{-1} \left( \mub_1 - \mub_2 \right) \left( \mub_1 - \mub_2 \right)^{T} \Gammab^{-1} \\
&= \underbrace{-\left( \Sigmab_{2} + \Lambda \right)^{-1} + \left( \Sigmab_{2} + \Lambda \right)^{-1} \Sigmab_1 \Gammab^{-1}}_{=-\Gammab^{-1}} + \Gammab^{-1} \left( \mub_1 - \mub_2 \right) \left( \mub_1 - \mub_2 \right)^{T} \Gammab^{-1} \\
\mathbf{R}_{12'\Sigmab} &= -\frac{1}{2}\left(\Gammab^{-1} - \gammab \otimes \gammab\right) \mathcal{N}(\mub_1; \mub_2, \Gammab)
\end{split}
\end{equation*}
\ref{eq:Ef}:
\begin{equation*}
\begin{split}
\mathbf{R}_{1'2'\mub\mub} &= \iint_{\X} \nabla_{\mub} \ln{\nu_{\theta}}|_{\theta_{t_1}}(\x) \otimes \nabla_{\mub} \ln{\nu_{\theta}}|_{\theta_{t_2}}(\x') \kb(\x, \x') d\nu_{\theta_{t_1}}(\x)d\nu_{\theta_{t_2}}(\x') \\
 &= \int_{\X} \nabla_{\mub} \ln{\nu_{\theta}}|_{\theta_{t_1}}(\x) \otimes \int_{\X} \nabla_{\mub} \ln{\nu_{\theta}}|_{\theta_{t_2}}(\x') \kb(\x, \x') \nu_{\theta_{t_2}}(\x') d \x' \nu_{\theta_{t_1}}(\x) d\x \\
 &= \Sigmab_{1}^{-1} \underbrace{\int_{\X} \left(\x - \mub_1 \right) \left(\x - \mub_2 \right)^{T} \mathcal{N}(\x; \mub_1,  \Sigmab_1) \mathcal{N}(\x; \mub_2,  \Sigmab_2 + \Lambda) d\x}_{(1)} \left( \Sigmab_{2} + \Lambda \right)^{-1}\\
 (1) &= \left[\Sigmab_1 \Gammab^{-1}\left(\Sigmab_2 + \Lambda \right) + \Sigmab_1 \Gammab^{-1} \left(\mub_1 - \mub_2 \right) \left(\mub_1 - \mub_2 \right)^{T} \Gammab^{-1} \Sigmab_1 - \Sigmab_1 \Gammab^{-1} \left(\mub_1 - \mub_2 \right) \left(\mub_1 - \mub_2 \right)^{T} \right] \\ 
 & \hspace{350pt} \times \mathcal{N}(\mub_1; \mub_2,  \Gammab) \\
 \mathbf{R}_{1'2'\mub\mub} &= \left[ \Gammab^{-1} + \underbrace{\Gammab^{-1} \left(\mub_1 - \mub_2 \right) \left(\mub_1 - \mub_2 \right)^{T} \Gammab^{-1} \Sigmab_1 \left( \Sigmab_2 + \Lambda \right)^{-1} - \Gammab^{-1} \left(\mub_1 - \mub_2 \right) \left(\mub_1 - \mub_2 \right)^{T} \left( \Sigmab_2 + \Lambda \right)^{-1}}_{= - \Gammab^{-1} \left(\mub_1 - \mub_2 \right) \left(\mub_1 - \mub_2 \right)^{T} \Gammab^{-1}} \right] \\
 & \hspace{350pt} \times \mathcal{N}(\mub_1; \mub_2,  \Gammab) \\
 &= \left[ \Gammab^{-1}-\Gammab^{-1} \left(\mub_1 - \mub_2 \right) \left(\mub_1 - \mub_2 \right)^{T} \Gammab^{-1}\right] \mathcal{N}(\mub_1; \mub_2, \Gammab)
\end{split}
\end{equation*}
\ref{eq:Eg}:
\begin{equation*}
\begin{split}
\mathbf{R}_{1'2'\mub\Sigmab} &= \iint_{\X} \nabla_{\mub} \ln{\nu_{\theta}}|_{\theta_{t_1}}(\x) \otimes \nabla_{\Sigmab} \ln{\nu_{\theta}}|_{\theta_{t_2}}(\x') \kb(\x, \x') d\nu_{\theta_{t_1}}(\x)d\nu_{\theta_{t_2}}(\x') \\
&= \int_{\X} \Sigmab_{1}^{-1} \left(\x - \mub_1 \right) \otimes \frac{1}{2} \left(\Lambda + \Sigmab_2\right)^{-1} \left[ \left(\x - \mub_2 \right) \left(\x - \mub_2 \right)^{T} - \left(\Lambda + \Sigmab_2\right) \right] \left(\Lambda + \Sigmab_2\right)^{-1} \\
& \hspace{250pt} \mathcal{N}(\x; \mub_1, \Sigmab_1)  \mathcal{N}(\x; \mub_2, \Sigmab_2 + \Lambda) d\x \\
&= \frac{1}{2} \Sigmab_{1}^{-1} *_{1} \left(\Lambda + \Sigmab_2\right)^{-1} *_{2} \left(\Lambda + \Sigmab_2\right)^{-1} *_{3}  \int_{\X} \left(\x - \mub_1 \right) \otimes \left(\x - \mub_2 \right) \otimes \left(\x - \mub_2 \right)\\
& \hspace{180pt} \underbrace{\hspace{70pt} \mathcal{N}(\x; \mub_1, \Sigmab_1)  \mathcal{N}(\x; \mub_2, \Sigmab_2 + \Lambda) d\x}_{(1)} \\
&- \frac{1}{2} \Sigmab_{1}^{-1} \underbrace{\int_{\X} \left(\x - \mub_1 \right) \mathcal{N}(\x; \mub_1, \Sigmab_1)  \mathcal{N}(\x; \mub_2, \Sigmab_2 + \Lambda) d\x}_{(2)} \otimes \left(\Lambda + \Sigmab_2\right)^{-1}\\
(1) &= \int_{\X} \x \otimes \x \otimes \x \mathcal{N}(\x; \mub_1 - \mub_2, \Sigmab_1) \mathcal{N}(\x; \boldsymbol{0}, \Sigmab_2 + \Lambda) d\x\\
&\hspace{50pt} -\left(\mub_1 - \mub_2 \right) \otimes \int_{\X} \x \otimes \x \mathcal{N}(\x; \mub_1 - \mub_2, \Sigmab_1 )\mathcal{N}(\x; \boldsymbol{0}, \Sigmab_2 + \Lambda) d\x \\
&= \mathcal{N}(\mub_1; \mub_2, \Gammab) \left( \nub \otimes \nub \otimes \nub + \sum\limits_{\tau \in \Taub_3} \tau(\nub \otimes \Pib) -\left(\mub_1 - \mub_2 \right) \otimes \left(\Pib + \nub \otimes \nub \right) \right) \\
& \text{ with } \left\{
    \begin{array}{ll}
        \Pib = \Sigmab_{1} \Gammab^{-1}\left( \Sigmab_2 + \Lambda \right) \\
        \nub = \Pib \Sigmab_{1}^{-1}\left( \mub_{1} - \mub_{2}\right)
    \end{array}
\right. \\
(2) &= \Sigmab_1 \Gammab^{-1} \left(\mub_2 - \mub_1 \right) \mathcal{N}(\mub_1; \mub_2, \Gammab)\\
\mathbf{R}_{1'2'\mub\Sigmab} &= \frac{1}{2} \Sigmab_{1}^{-1} *_{1} \left(\Lambda + \Sigmab_2\right)^{-1} *_{2} \left(\Lambda + \Sigmab_2\right)^{-1} *_{3} (1) - \frac{1}{2} \Sigmab_{1}^{-1} (2) \otimes  \left(\Lambda + \Sigmab_2\right)^{-1} \\
&= \frac{1}{2} \mathcal{N}(\mub_1; \mub_2, \Gammab) \Bigg \{ \\
(a) \quad & \Sigmab_{1}^{-1} \left(\Lambda + \Sigmab_2\right) \Gammab^{-1} \left(\mub_1 - \mub_2 \right) \otimes \Gammab^{-1} \left(\mub_1 - \mub_2 \right) \otimes \Gammab^{-1} \left(\mub_1 - \mub_2 \right) \\
(b) \quad &+ \Sigmab_{1}^{-1} \left(\Lambda + \Sigmab_2\right) \Gammab^{-1} \left(\mub_1 - \mub_2 \right) \otimes \left(\Lambda + \Sigmab_2\right)^{-1} \Sigmab_{1} \Gammab^{-1}\\
(c) \quad &+ \sigma \left( \Gammab^{-1} \left(\mub_1 - \mub_2 \right) \otimes \Gammab^{-1}\right)\\
(d) \quad &+ \Gammab^{-1} \otimes \Gammab^{-1} \left(\mub_1 - \mub_2 \right)\\
(e) \quad &- \Sigmab_{1}^{-1} \left(\mub_1 - \mub_2 \right) \otimes \left( \Lambda + \Sigmab_2 \right)^{-1} \Sigmab_{1} \Gammab^{-1}\\
(f) \quad &- \Sigmab_{1}^{-1} \left(\mub_1 - \mub_2 \right) \otimes \Gammab^{-1} \left(\mub_1 - \mub_2 \right) \otimes \Gammab^{-1} \left(\mub_1 - \mub_2 \right)\\
(g) \quad &- \Sigmab_{1}^{-1} \left(\mub_2 - \mub_1 \right) \otimes \left(\Lambda + \Sigmab_2\right)^{-1} \Bigg\}\\
(a) + (f) &= \left[\Sigmab_{1}^{-1} \left(\Lambda + \Sigmab_2\right) \Gammab^{-1} - \Sigmab_{1}^{-1} \right] \left(\mub_1 - \mub_2 \right) \otimes \Gammab^{-1} \left(\mub_1 - \mub_2 \right) \otimes \Gammab^{-1} \left(\mub_1 - \mub_2 \right)\\
&= -\Gammab^{-1} \left(\mub_1 - \mub_2 \right) \otimes \Gammab^{-1} \left(\mub_1 - \mub_2 \right) \otimes \Gammab^{-1} \left(\mub_1 - \mub_2 \right)\\
(b) + (e) &= \left[\Sigmab_{1}^{-1} \left(\Lambda + \Sigmab_2\right) \Gammab^{-1} - \Sigmab_{1}^{-1} \right]\left(\mub_1 - \mub_2 \right) \otimes \left(\Lambda + \Sigmab_2\right)^{-1} \Sigmab_{1} \Gammab^{-1}\\
&= -\Gammab^{-1}\left(\mub_1 - \mub_2\right) \otimes \left(\Lambda + \Sigmab_2\right)^{-1} \Sigmab_{1} \Gammab^{-1}\\
(b) + (e) + (g)&= -\Gammab^{-1}\left(\mub_1 - \mub_2\right) \otimes \left[ \left(\Lambda + \Sigmab_2\right)^{-1} \Sigmab_{1} \Gammab^{-1} - \left(\Lambda + \Sigmab_2\right)^{-1} \right]\\
&= \Gammab^{-1}\left(\mub_1 - \mub_2\right) \otimes \Gammab^{-1}\\
\mathbf{R}_{1'2'\mub\Sigmab} &= \frac{1}{2} \left( \sum\limits_{\tau \in \Taub_3} \tau(\gammab \otimes \Gammab^{-1}) - \gammab \otimes \gammab \otimes \gammab \right)\mathcal{N}(\mub_1; \mub_2, \Gammab)
\end{split}
\end{equation*}
\ref{eq:Eh}:
\begin{equation*}
\begin{split}
\mathbf{R}_{1'2'\Sigmab\Sigmab} &= \iint_{\X} \nabla_{\Sigmab} \ln{\nu_{\theta}}|_{\theta_{t_1}}(\x) \otimes \nabla_{\Sigmab} \ln{\nu_{\theta}}|_{\theta_{t_2}}(\x') \kb(\x, \x') d\nu_{\theta_{t_1}}(\x)d\nu_{\theta_{t_2}}(\x') \\
&= \int_{\X} \frac{1}{2}\big[-\Sigmab_{1}^{-1} +\Sigmab_{1}^{-1}(\x - \mub_1)(\x - \mub_1)^{T}\Sigmab_{1}^{-1} \big] \otimes \frac{1}{2} \left(\Lambda + \Sigmab_2\right)^{-1} \left[ \left(\x - \mub_2 \right) \left(\x - \mub_2 \right)^{T} - \left(\Lambda + \Sigmab_2\right) \right] \left(\Lambda + \Sigmab_2\right)^{-1} \\
& \hspace{250pt} \mathcal{N}(\x; \mub_1, \Sigmab_1)  \mathcal{N}(\x; \mub_2, \Sigmab_2 + \Lambda) d\x \\
&= \frac{1}{4} \int_{\X} \Bigg\{ \underbrace{\Sigmab_{1}^{-1} \otimes \left(\Lambda + \Sigmab_2\right)^{-1}}_{(1)} \\
&- \Sigmab_{1}^{-1} \otimes \left(\Lambda + \Sigmab_2\right)^{-1} \left(\x - \mub_2 \right) \otimes \left(\Lambda + \Sigmab_2\right)^{-1} \left(\x - \mub_2 \right) \quad (2) \\
&+ \Sigmab_{1}^{-1} \left(\x - \mub_1 \right) \otimes \Sigmab_{1}^{-1} \left(\x - \mub_1 \right) \otimes \left(\Lambda + \Sigmab_2\right)^{-1} \left(\x - \mub_2 \right) \otimes \left(\Lambda + \Sigmab_2\right)^{-1} \left(\x - \mub_2 \right) \quad (3) \\
&- \Sigmab_{1}^{-1} \left(\x - \mub_1 \right) \otimes \Sigmab_{1}^{-1} \left(\x - \mub_1 \right) \otimes \left(\Lambda + \Sigmab_2\right)^{-1} \quad (4) \\
& \Bigg\} \mathcal{N}(\x; \mub_1, \Sigmab_1)  \mathcal{N}(\x; \mub_2, \Sigmab_2 + \Lambda) d\x \\
(1) &: \int_{\X} \Sigmab_{1}^{-1} \otimes \left(\Lambda + \Sigmab_2\right)^{-1} \mathcal{N}(\x; \mub_1, \Sigmab_1)  \mathcal{N}(\x; \mub_2, \Sigmab_2 + \Lambda) d\x \\
&= \Sigmab_{1}^{-1} \otimes \left(\Lambda + \Sigmab_2\right)^{-1} \mathcal{N}(\mub_1; \mub_2, \Sigmab_1 + \Sigmab_2 + \Lambda) \\
(2) &: -\int_{\X} \Sigmab_{1}^{-1} \otimes \left(\Lambda + \Sigmab_2\right)^{-1} \left(\x - \mub_2 \right) \otimes \left(\Lambda + \Sigmab_2\right)^{-1} \left(\x - \mub_2 \right) \mathcal{N}(\x; \mub_1, \Sigmab_1)  \mathcal{N}(\x; \mub_2, \Sigmab_2 + \Lambda) d\x \\
&= -\Sigmab_{1}^{-1} \otimes \left(\Lambda + \Sigmab_2\right)^{-1} *_{1} \left(\Lambda + \Sigmab_2\right)^{-1} *_{2} \left(\Pib + \nub \otimes \nub \right) \mathcal{N}(\mub_1; \mub_2, \Sigmab_1 + \Sigmab_2 + \Lambda) \\
& \text{ with } \left\{
    \begin{array}{ll}
        \Pib = \Sigmab_{1} \Gammab^{-1}\left( \Sigmab_2 + \Lambda \right) \\
        \nub = \Pib \Sigmab_{1}^{-1}\left( \mub_{1} - \mub_{2}\right)
    \end{array}
\right. \\
&= -\Sigmab_{1}^{-1} \otimes \left(\Lambda + \Sigmab_2\right)^{-1} *_{1} \left(\Lambda + \Sigmab_2\right)^{-1} *_{2} \big( \Sigmab_{1} \Gammab^{-1}\left( \Sigmab_2 + \Lambda \right) \\
& \quad +  \left(\Lambda + \Sigmab_2\right) \Gammab^{-1} \left( \mub_{1} - \mub_{2}\right) \otimes \left(\Lambda + \Sigmab_2\right) \Gammab^{-1} \left( \mub_{1} - \mub_{2}\right) \big) \mathcal{N}(\mub_1; \mub_2, \Sigmab_1 + \Sigmab_2 + \Lambda) \\
&= -\Sigmab_{1}^{-1} \otimes \left\{ \Gammab^{-1} \Sigmab_{1} \left(\Sigmab_2 +\Lambda \right)^{-1} + \Gammab^{-1} \left( \mub_{1} - \mub_{2}\right) \otimes \Gammab^{-1} \left( \mub_{1} - \mub_{2}\right)  \right\} \mathcal{N}(\mub_1; \mub_2, \Gammab) \\
(4) &: -\int_{\X} \Sigmab_{1}^{-1} \left(\x - \mub_1 \right) \otimes \Sigmab_{1}^{-1} \left(\x - \mub_1 \right) \otimes \left(\Sigmab_2 +\Lambda \right)^{-1} \mathcal{N}(\x; \mub_1, \Sigmab_1)  \mathcal{N}(\x; \mub_2, \Sigmab_2 + \Lambda) d\x \\
&= - \Sigmab_{1}^{-1} *_{1} \Sigmab_{1}^{-1} *_{2} \left(\Pib + \nub \otimes \nub \right) \otimes \left(\Sigmab_2 +\Lambda \right)^{-1} \mathcal{N}(\mub_1; \mub_2, \Gammab) \\
& \text{ with } \left\{
    \begin{array}{ll}
        \Pib = \Sigmab_{1} \Gammab^{-1}\left( \Sigmab_2 + \Lambda \right) \\
        \nub = \Pib \left(\Sigmab_2 +\Lambda \right)^{-1}\left( \mub_{2} - \mub_{1}\right)
    \end{array}
\right. \\
&= - \Sigmab_{1}^{-1} *_{1} \Sigmab_{1}^{-1} *_{2} \left( \Sigmab_{1} \Gammab^{-1}\left( \Sigmab_2 + \Lambda \right)  + \Sigmab_{1} \Gammab^{-1} \left( \mub_{2} - \mub_{1}\right)  \otimes \Sigmab_{1} \Gammab^{-1} \left( \mub_{2} - \mub_{1}\right) \right) \otimes \left(\Sigmab_2 +\Lambda \right)^{-1} \mathcal{N}(\mub_1; \mub_2, \Gammab) \\
&= - \left[ \Gammab^{-1}\left( \Sigmab_2 + \Lambda \right) \Sigmab_{1}^{-1} + \Gammab^{-1} \left( \mub_{2} - \mub_{1}\right)  \otimes \Gammab^{-1} \left( \mub_{2} - \mub_{1}\right) \right] \otimes \left( \Sigmab_2 + \Lambda \right)^{-1} \mathcal{N}(\mub_1; \mub_2, \Gammab) \\
(3) &: \int_{\X} \Sigmab_{1}^{-1} \left(\x - \mub_1 \right) \otimes \Sigmab_{1}^{-1} \left(\x - \mub_1 \right) \otimes \left( \Sigmab_2 + \Lambda \right)^{-1} \left(\x - \mub_2 \right) \otimes \left( \Sigmab_2 + \Lambda \right)^{-1} \left(\x - \mub_2 \right) \\
& \hspace{250pt} \mathcal{N}(\x; \mub_1, \Sigmab_1)  \mathcal{N}(\x; \mub_2, \Sigmab_2 + \Lambda) d\x \\
&= \Sigmab_{1}^{-1} *_{1} \Sigmab_{1}^{-1} *_{2} \left( \Sigmab_2 + \Lambda \right)^{-1} *_{3} \left( \Sigmab_2 + \Lambda \right)^{-1} *_{4} \int_{\X} \left(\x - \mub_1 \right) \otimes \left(\x - \mub_1 \right) \otimes \left(\x - \mub_2 \right) \otimes \left(\x - \mub_2 \right) \\
& \hspace{210pt} \underbrace{\hspace{40pt} \mathcal{N}(\x; \mub_1, \Sigmab_1)  \mathcal{N}(\x; \mub_2, \Sigmab_2 + \Lambda) d\x}_{(*)} \\
(*) &= \int_{\X}  \left(\left(\x - \nub \right) + \left(\nub - \mub_1 \right) \right) \otimes \left(\left(\x - \nub \right) + \left(\nub - \mub_1 \right) \right) \otimes \left(\left(\x - \nub \right) + \left(\nub - \mub_2 \right) \right) \otimes \left(\left(\x - \nub \right) + \left(\nub - \mub_2 \right) \right) \\
& \hspace{250pt} \mathcal{N}(\x; \nub, \Pib) d\x \mathcal{N}(\mub_1; \mub_2, \Gammab) \\
& \text{ with } \left\{
    \begin{array}{ll}
        \Pib = \Sigmab_{1} \Gammab^{-1}\left( \Sigmab_2 + \Lambda \right) \\
        \nub = \Pib \Sigmab_1^{-1} \mub_{1} + \Pib \left(\Sigmab_2 +\Lambda \right)^{-1} \mub_{2} = \left(\Sigmab_2 +\Lambda \right) \Gammab^{-1} \mub_{1} + \Sigmab_1^{-1} \Gammab^{-1} \mub_{2}
    \end{array}
\right. \\
\end{split}
\end{equation*}
\begin{equation*}
\begin{split}
(*) &= \mathcal{N}(\mub_1; \mub_2, \Gammab) \int_{\X} \Big[ \underbrace{\left(\x - \nub \right) \otimes \left(\x - \nub \right) \otimes\left(\x - \nub \right) \otimes\left(\x - \nub \right)}_{(a)}\\
& \left.
    \begin{array}{llll}
        + \left(\nub - \mub_1 \right) \otimes \left(\x - \nub \right) \otimes\left(\x - \nub \right) \otimes\left(\x - \nub \right) \\
        + \left(\x - \nub \right) \otimes \left(\nub - \mub_1 \right) \otimes \left(\x - \nub \right) \otimes\left(\x - \nub \right) \\
        + \left(\x - \nub \right) \otimes \left(\x - \nub \right) \otimes\left(\nub - \mub_2 \right) \otimes\left(\x - \nub \right) \\
        + \left(\x - \nub \right) \otimes \left(\x - \nub \right) \otimes \left(\x - \nub \right) \otimes \left(\nub - \mub_2 \right) 
    \end{array}
\right\} \quad (b) \\
& \left.
    \begin{array}{llllll}
        + \left(\x - \nub \right) \otimes \left(\x - \nub \right) \otimes\left(\nub - \mub_2 \right) \otimes\left(\nub - \mub_2 \right) \\
        + \left(\x - \nub \right) \otimes \left(\nub - \mub_1 \right) \otimes \left(\x - \nub \right) \otimes\left(\nub - \mub_2 \right) \\
        + \left(\x - \nub \right) \otimes \left(\nub - \mub_1 \right) \otimes \left(\nub - \mub_2 \right) \otimes \left(\x - \nub \right) \\
        + \left(\nub - \mub_1 \right) \otimes \left(\x - \nub \right) \otimes \left(\x - \nub \right) \otimes \left(\nub - \mub_2 \right) \\
        + \left(\nub - \mub_1 \right) \otimes \left(\x - \nub \right) \otimes\left(\nub - \mub_2 \right) \otimes\left(\x - \nub \right) \\
        + \left(\nub - \mub_1 \right) \otimes \left(\nub - \mub_1 \right) \otimes \left(\x - \nub \right) \otimes \left(\x - \nub \right) 
    \end{array}
\right\} \quad (c) \\
& \left.
    \begin{array}{llll}
        + \left(\x - \nub \right) \otimes \left(\nub - \mub_1 \right) \otimes \left(\nub - \mub_2 \right) \otimes \left(\nub - \mub_2 \right) \\
        + \left(\nub - \mub_1 \right) \otimes \left(\x - \nub \right) \otimes \left(\nub - \mub_2 \right) \otimes \left(\nub - \mub_2 \right) \\
        + \left(\nub - \mub_1 \right) \otimes \left(\nub - \mub_1 \right) \otimes \left(\x - \nub \right) \otimes \left(\nub - \mub_2 \right) \\
        + \left(\nub - \mub_1 \right) \otimes \left(\nub - \mub_1 \right) \otimes \left(\nub - \mub_2 \right) \otimes \left(\x - \nub \right)
    \end{array}
\right\} \quad (d) \\
&+ \underbrace{\left(\nub - \mub_1 \right) \otimes \left(\nub - \mub_1 \right) \otimes \left(\nub - \mub_2 \right) \otimes \left(\nub - \mub_2 \right)}_{(e)} \Big] \mathcal{N}(\x; \nub, \Pib) d\x \\
(a) &: \int_{\X} \left(\x - \nub \right) \otimes \left(\x - \nub \right) \otimes\left(\x - \nub \right) \otimes\left(\x - \nub \right) \mathcal{N}(\x; \nub, \Pib) d\x \\
&= \Pib \otimes \Pib + \left[\Pib \otimes \Pib\right]_{1324} + \left[\Pib \otimes \Pib\right]_{1423} \text{ where } \left[ \left[\Sigmab\right]_{i_1 i_2 i_3 i_4}\right]_{j_1 j_2 j_3 j_4} = \left[\Sigmab\right]_{j_{i_1} j_{i_2} j_{i_3} j_{i_4}}\\
(b) &: \int_{\X} (b) \mathcal{N}(\x; \nub, \Pib) d\x = 0\\
(c) &: \int_{\X} (c) \mathcal{N}(\x; \nub, \Pib) d\x = \Pib \otimes \left(\nub - \mub_2 \right) \otimes \left(\nub - \mub_2 \right) + \left(\nub - \mub_1 \right) \otimes \left(\nub - \mub_1 \right) \otimes \Pib \\
&+ \left(\nub - \mub_1 \right) \otimes \Pib \otimes \left(\nub - \mub_2 \right) + \left[\left(\nub - \mub_1 \right) \otimes \Pib \otimes \left(\nub - \mub_2 \right)\right]_{2134} + \left[\left(\nub - \mub_1 \right) \otimes \Pib \otimes \left(\nub - \mub_2 \right)\right]_{1243} \\
& \hspace{280pt} + \left[\left(\nub - \mub_1 \right) \otimes \Pib \otimes \left(\nub - \mub_2 \right)\right]_{2143}\\
(d) &: \int_{\X} (d) \mathcal{N}(\x; \nub, \Pib) d\x = 0\\
(e) &: \int_{\X} \left(\nub - \mub_1 \right) \otimes \left(\nub - \mub_1 \right) \otimes \left(\nub - \mub_2 \right) \otimes \left(\nub - \mub_2 \right) \mathcal{N}(\x; \nub, \Pib) d\x\\
&= \left(\nub - \mub_1 \right) \otimes \left(\nub - \mub_1 \right) \otimes \left(\nub - \mub_2 \right) \otimes \left(\nub - \mub_2 \right) \\
(3) &= \Sigmab_{1}^{-1} *_{1} \Sigmab_{1}^{-1} *_{2} \left( \Sigmab_2 + \Lambda \right)^{-1} *_{3} \left( \Sigmab_2 + \Lambda \right)^{-1} *_{4} (*) \\
& \text{We denote (3a):(3e) this multiplication with the terms (a):(e) respectively.} \\
(3a) &= \Gammab^{-1} \left( \Sigmab_2 + \Lambda \right) \Sigmab_{1}^{-1} \otimes \Gammab^{-1} \Sigmab_{1} \left( \Sigmab_2 + \Lambda \right)^{-1} + \left[\Gammab^{-1} \otimes \Gammab^{-1}\right]_{1324} + \left[\Gammab^{-1} \otimes \Gammab^{-1}\right]_{1423}\\
(3c) &= \Gammab^{-1} \left( \Sigmab_2 + \Lambda \right) \Sigmab_{1}^{-1} \otimes \Gammab^{-1} \left(\mub_1 - \mub_2 \right) \otimes \Gammab^{-1} \left(\mub_1 - \mub_2 \right) \\
&+\Gammab^{-1} \left(\mub_1 - \mub_2 \right) \otimes \Gammab^{-1} \left(\mub_1 - \mub_2 \right) \otimes  \Gammab^{-1} \Sigmab_{1} \left( \Sigmab_2 + \Lambda \right)^{-1} \\
&+ \Gammab^{-1} \left(\mub_2 - \mub_1 \right) \otimes \Gammab^{-1} \otimes \Gammab^{-1} \left(\mub_1 - \mub_2 \right) + \left[ \Gammab^{-1} \left(\mub_2 - \mub_1 \right) \otimes \Gammab^{-1} \otimes \Gammab^{-1} \left(\mub_1 - \mub_2 \right) \right]_{2134} \\
&+ \left[ \Gammab^{-1} \left(\mub_2 - \mub_1 \right) \otimes \Gammab^{-1} \otimes \Gammab^{-1} \left(\mub_1 - \mub_2 \right) \right]_{1243} + \left[ \Gammab^{-1} \left(\mub_2 - \mub_1 \right) \otimes \Gammab^{-1} \otimes \Gammab^{-1} \left(\mub_1 - \mub_2 \right) \right]_{2143} \\
(3e) &= \Gammab^{-1}  \left(\mub_2 - \mub_1 \right) \otimes \Gammab^{-1}  \left(\mub_2 - \mub_1 \right) \otimes \Gammab^{-1}  \left(\mub_1 - \mub_2 \right) \otimes \Gammab^{-1}  \left(\mub_1 - \mub_2 \right)\\
\end{split}
\end{equation*}
\begin{equation*}
\begin{split}
\mathbf{R}_{1'2'\Sigmab\Sigmab} &= \frac{1}{4} \mathcal{N}(\mub_1; \mub_1, \Gammab) \Bigg\{ \underbrace{\Sigmab_{1}^{-1} \otimes \left(\Sigmab_{2} + \Lambda \right)^{-1}}_{(1a)}\\
&\underbrace{- \Sigmab_{1}^{-1} \otimes \Gammab^{-1} \Sigmab_{1} \left(\Sigmab_2 +\Lambda \right)^{-1}}_{(1b)} \underbrace{- \Sigmab_{1}^{-1} \otimes \Gammab^{-1} \left( \mub_{1} - \mub_{2}\right) \otimes \Gammab^{-1} \left( \mub_{1} - \mub_{2}\right)}_{(2a)} \\
&\underbrace{+ \Gammab^{-1} \left( \Sigmab_2 + \Lambda \right) \Sigmab_{1}^{-1} \otimes \Gammab^{-1} \Sigmab_{1} \left( \Sigmab_2 + \Lambda \right)^{-1}}_{(1c)} + \left[\Gammab^{-1} \otimes \Gammab^{-1}\right]_{1324} + \left[\Gammab^{-1} \otimes \Gammab^{-1}\right]_{1423} \\
&\underbrace{+ \Gammab^{-1} \left( \Sigmab_2 + \Lambda \right) \Sigmab_{1}^{-1} \otimes \Gammab^{-1} \left(\mub_1 - \mub_2 \right) \otimes \Gammab^{-1} \left(\mub_1 - \mub_2 \right)}_{(2b)} \\
&\underbrace{+\Gammab^{-1} \left(\mub_1 - \mub_2 \right) \otimes \Gammab^{-1} \left(\mub_1 - \mub_2 \right) \otimes  \Gammab^{-1} \Sigmab_{1} \left( \Sigmab_2 + \Lambda \right)^{-1}}_{(2c)} \\
&+ \Gammab^{-1} \left(\mub_2 - \mub_1 \right) \otimes \Gammab^{-1} \otimes \Gammab^{-1} \left(\mub_1 - \mub_2 \right) + \left[ \Gammab^{-1} \left(\mub_2 - \mub_1 \right) \otimes \Gammab^{-1} \otimes \Gammab^{-1} \left(\mub_1 - \mub_2 \right) \right]_{2134} \\
&+ \left[ \Gammab^{-1} \left(\mub_2 - \mub_1 \right) \otimes \Gammab^{-1} \otimes \Gammab^{-1} \left(\mub_1 - \mub_2 \right) \right]_{1243} + \left[ \Gammab^{-1} \left(\mub_2 - \mub_1 \right) \otimes \Gammab^{-1} \otimes \Gammab^{-1} \left(\mub_1 - \mub_2 \right) \right]_{2143}\\
&+ \Gammab^{-1}  \left(\mub_2 - \mub_1 \right) \otimes \Gammab^{-1}  \left(\mub_2 - \mub_1 \right) \otimes \Gammab^{-1}  \left(\mub_1 - \mub_2 \right) \otimes \Gammab^{-1}  \left(\mub_1 - \mub_2 \right)\\
&\underbrace{- \Gammab^{-1}\left( \Sigmab_2 + \Lambda \right) \Sigmab_{1}^{-1} \otimes \left( \Sigmab_2 + \Lambda \right)^{-1}}_{(1d)} \underbrace{- \Gammab^{-1} \left( \mub_{2} - \mub_{1}\right)  \otimes \Gammab^{-1} \left( \mub_{2} - \mub_{1}\right) \otimes \left( \Sigmab_2 + \Lambda \right)^{-1}}_{(2d)} \Bigg\}
\end{split}
\end{equation*}
We have:
\begin{equation*}
\begin{split}
    (1a) + (1b) + (1c) + (1d) &= \Gammab^{-1} \otimes \Gammab^{-1} \\
    (2a) + (2b) &= - \Gammab^{-1} \otimes \Gammab^{-1} \left( \mub_{2} - \mub_{1}\right)  \otimes \Gammab^{-1} \left( \mub_{2} - \mub_{1}\right) \\
    (2c) + (2d) &= - \Gammab^{-1} \left( \mub_{2} - \mub_{1}\right)  \otimes \Gammab^{-1} \left( \mub_{2} - \mub_{1}\right) \otimes \Gammab^{-1}
\end{split}
\end{equation*}
We define 
\ref{eq:Ei}:
Immediate from \ref{eq:productint}.\\
\ref{eq:Ej}:
\begin{equation*}
\begin{split}
\left[\boldsymbol{\mathcal{T}}_{kX}\right]_{i, \mub} &= \int_{\X} \Sigmab^{-1}_{k} \left( \x - \mub_{k} \right) \mathcal{N}(\x; \x_i, \Lambda) \mathcal{N}(\x; \mub_{k}, \Sigmab_{k}) d\x \\
&= \left(\Sigmab_{k} + \Lambda \right)^{-1} \left( \x_i - \mub_{k} \right) \mathcal{N}(\x_i; \mub_{k}, \Sigmab_{k} + \Lambda)
\end{split}
\end{equation*}
\ref{eq:Ek}:
\begin{equation*}
\begin{split}
\left[\boldsymbol{\mathcal{T}}_{kX}\right]_{i, \Sigmab} &= \int_{\X} \frac{1}{2}\big[-\Sigmab^{-1}_{k} +\Sigmab^{-1}_{k}(\x - \mub_{k})(\x - \mub_{k})^{T}\Sigmab^{-1}_{k} \big] \mathcal{N}(\x; \x_i, \Lambda) \mathcal{N}(\x; \mub_{k}, \Sigmab_{k}) d\x \\
&= -\frac{1}{2}\Sigmab^{-1}_{k} \mathcal{N}(\x_i; \mub_{k}, \Sigmab_{k} + \Lambda) + \frac{1}{2} \Sigmab^{-1}_{k} \left[ \int_{\X} \x \otimes \x \mathcal{N}(\x; \x_i - \mub_{k}, \Lambda) \mathcal{N}(\x; \boldsymbol{0}, \Sigmab_{k} + \Lambda) d\x \right] \Sigmab^{-1}_{k}\\
&= \frac{1}{2} \left[-\Sigmab^{-1}_{k} + \left(\Sigmab_{k} + \Lambda \right)^{-1} \Lambda \Sigmab^{-1}_{k} + \left(\Sigmab_{k} + \Lambda \right)^{-1} \left(\x_i - \mub_{k} \right) \left(\x_i - \mub_{k} \right)^{T} \left(\Sigmab_{k} + \Lambda \right)^{-1} \right] \mathcal{N}(\x_i; \mub_{k}, \Sigmab_{k} + \Lambda) \\
&= \frac{1}{2} \left[- \left(\Sigmab_{k} + \Lambda \right)^{-1} + \left(\Sigmab_{k} + \Lambda \right)^{-1} \left(\x_i - \mub_{k} \right) \left(\x_i - \mub_{k} \right)^{T} \left(\Sigmab_{k} + \Lambda \right)^{-1} \right] \mathcal{N}(\x_i; \mub_{k}, \Sigmab_{k} + \Lambda)
\end{split}
\end{equation*}
    % \mathbf{R}_{1'2'} &= \iint_{\X} \nabla_{\theta} \ln{\nu_{\theta}}|_{\theta_{t_1}}(\x) \otimes \nabla_{\theta} \ln{\nu_{\theta}}|_{\theta_{t_2}}(\x') \kb(\x, \x') d\nu_{\theta_{t_1}}(\x)d\nu_{\theta_{t_2}}(\x') \\
    % \left[\mathbf{t}_{kX}\right]_{i} &= \int_{\X} \kb(\x, \x_i)d\nu_{\theta_k}(\x) \\
    % \left[\boldsymbol{\mathcal{T}}_{kX}\right]_{ij} &= \int_{\X} \left(\nabla_{\theta} \ln{\nu_{\theta}}|_{\theta_{t_k}}\right)_{j}(\x) \kb(\x, \x_i)d\nu_{\theta_{t_k}}(\x)
\end{proof}

\section{Natural Gradient} \label{appendix:naturalgradient}
If $\theta$, $\theta^{'} \in \Theta$ then we can define a distance over distributions as:
\begin{equation}
    KL(P_{\theta^{'}} || P_{\theta}) = \int_x \ln{\frac{P_{\theta^{'}}(dx)}{P_{\theta}(dx)}}P_{\theta^{'}}(dx)
\end{equation} Its expansion defines the Fisher Information matrix:
\begin{equation}
    KL(P_{\theta + \delta \theta} || P_{\theta}) = \frac{1}{2}\sum \boldsymbol{F}_{ij}(\theta) \delta \theta_i \delta \theta_j + O(\delta \theta^{3})
\end{equation}
\begin{equation} \label{eq:fisher}
    \boldsymbol{F}_{ij}(\theta) = -\int \frac{\partial^{2} \ln{P_{\theta}(x)}}{\partial \theta_i \partial \theta_j} P_{\theta}(dx)
\end{equation}
If $g : \Theta \rightarrow \mathbb{R}$ is a smooth function pf the parameter space then its natural gradient is:
\begin{equation}
    \widetilde{\nabla}_{\theta} g = \boldsymbol{F}^{-1} \frac{\partial g(\theta)}{\partial \theta}
\end{equation}
By construction, the natural gradient descent is intrinsic: it does not depend on the chosen parametrization $\theta$ of $P_{\theta}$. This perspective has been studied in the context of optimisation \cite{ollivier2017information}. The resulting algorithm, called IGO (Information Geometric Optimisation) exploit parameterization invariance via natural gradient descent and function invariance to monotonic transformation via rank comparison. When \cite{ollivier2017information} seek monoticity invariance by replacing $f$ in \ref{eq:parame} we will instead leverage bayesian modelling via Gaussian Processes to leverage function prior information. 

\textbf{Multivariate Gaussian Distribution}

In the case of multivariate Gaussian distributions $\mathcal{N}\left(\mub(\theta), \Sigmab(\theta) \right)$, the Fisher Information Matrix has the form:
\begin{equation*}
    \boldsymbol{F}_{i, j} = \frac{\partial\mub^{T}}{\partial\theta_i} \Sigmab^{-1}\frac{\partial\mub}{\partial \theta_j} + \frac{1}{2}tr\left(\Sigmab^{-1} \frac{\partial\Sigmab}{\partial\theta_i} \Sigmab^{-1} \frac{\partial\Sigmab}{\partial\theta_j}\right)
\end{equation*}

Given the parameterisation $\theta = (\mub, \text{vech}(\Sigmab))$ where $\text{vech}(\Sigmab) = \left(\Sigmab_{1:n, 1}, \Sigmab_{1:n, 2}, \dots, \Sigmab_{1:n, n}\right)$, we have $\boldsymbol{F}\left(\theta\right) = \begin{bmatrix}
\Sigmab^{-1} & \mathbf{0} \\
\mathbf{0} & \frac{1}{2} \Sigmab^{-1} \otimes_{K} \Sigmab^{-1} 
\end{bmatrix}$ where $\otimes_{K}$ is the Kronecker product. Hence, we have $\boldsymbol{F}^{-1}\left(\theta\right) = \begin{bmatrix}
\Sigmab & \mathbf{0} \\
\mathbf{0} & 2 \Sigmab \otimes_{K} \Sigmab 
\end{bmatrix}$.

\section{Probabilistic Evolutionary Search Formula and Proofs}
\label{appendix:probes}

\paragraph{xNES.} 
The algorithm xNES \citep{wierstra2014natural} uses an `exponential' parametrization of the input distribtution. With $\Sigmab^{t} = \mathbf{A}^{t}\left(\mathbf{A}^{t}\right)^{T}$, the xNES updates are:
\begin{equation*}
\begin{split}
    \mub^{t+1} &= \mub^{t} + \eta_{m}\sum\limits_{i=1}^{n} w_i \left( \x_i - \mub^t\right) \\
    \A^{t+1} &= \A^{t} \exp \Bigg(\frac{\eta}{2} \\
    & \sum\limits_{i=1}^{n} w_i \left(\left(\mathbf{A}^{t}\right)^{-1} \left( \x_i - \mub^t \right) \left( \x_i - \mub^t \right)^{T} \left(\left(\mathbf{A}^{t}\right)^{-1}\right)^{T} - \I \right)\Bigg)
\end{split}
\end{equation*}

\begin{prop}[\textbf{Probabilistic xNES}]
The update for the probabilistic xNES is:
\begin{equation}
\begin{split}
    \mub^{t+1} &= \mub^{t} + \eta \Sigmab\left(\Sigmab + \Lambda\right)^{-1}\sum\limits_{i=1}^{N} w_i \left( \x_i - \mub^t\right) \\
    \R^{t+1} &= \R^{t} + \eta \left(\mathbf{A}^{t}\right)^{T}\left(\Sigmab^{t} + \Lambda\right)^{-1} \Bigg[\sum\limits_{i=1}^{N}w_i \\
    & \left(\left( \x_i - \mub^t \right) \left( \x_i - \mub^t \right)^{T} - \left(\Sigmab^{t} + \Lambda\right)\right) \Bigg] \left(\Sigmab^{t} + \Lambda\right)^{-1}\A^{t} \\
    \R^{t} &= \ln \left( \left(\mathbf{A}^{t}\right)^{-1} \Sigmab^{t} \left( \left(\mathbf{A}^{t}\right)^{T} \right)^{-1}\right)\\
    \mathbf{w} &= \left[\kb_{\Xb\Xb} + \sigma_{noise}^{2}\mathbf{I}\right]^{-1} (\y  - \mb_{\Xb}) \odot \mathcal{N}\left(\Xb; \mub^{t}, \Lambda + \Sigmab^{t}\right)
\end{split}
\end{equation}
\end{prop}
\begin{proof}
We follow the argument given in the proof of Proposition 16 in \citet{ollivier2017information}.
\end{proof}

\paragraph{SNES.} 
The algorithm Separable NES (SNES; \citet{wierstra2014natural}) assumes the search distribution is given by a multivariate normal distribution with diagonal covariance matrix and is more suitable for high dimensional search space. With $\Sigmab^{t} = Diag(\boldsymbol{\sigma}_{t}^{2})$, the SNES updates are:
\begin{equation*}
\begin{split}
    \mub^{t+1} &= \mub^{t} + \eta\sum\limits_{i=1}^{n} w_i \left( \x_i - \mub^t\right) \\
    \boldsymbol{\sigma}_{t+1} &= \boldsymbol{\sigma}_{t} \exp \left(\frac{\eta}{2} \sum\limits_{i=1}^{n} w_i \left(\left( \left(\boldsymbol{\sigma}_{t}\right)^{-1} \left( \x_i - \mub^t \right) \right)^{2} - \I \right)\right)
\end{split}
\end{equation*}

\begin{prop}[\textbf{Probabilistic SNES}]
Let $\Lambda = Diag(\boldsymbol{\lambda}^{2})$ where $\boldsymbol{\lambda} = [\lambda_1, ..., \lambda_D]$. The update for the probabilistic SNES is:
\begin{equation}
\begin{split}
    \mub^{t+1} &= \mub^{t} + \eta \frac{\boldsymbol{\sigma}_{t}^{2}}{\boldsymbol{\lambda}^{2} + \boldsymbol{\sigma}_{t}^{2}}\sum\limits_{i=1}^{N} w_i \left( \x_i - \mub^t\right) \\
    \boldsymbol{\sigma}_{t+1} &= \boldsymbol{\sigma}_{t} \exp \Bigg(\\
    & \hspace{35pt}\frac{\eta}{2}\frac{\boldsymbol{\sigma}_{t}^{2}}{\boldsymbol{\sigma}_{t}^{2} + \boldsymbol{\lambda}_{t}^{2}} \sum\limits_{i=1}^{n} w_i \left( \left( \frac{\x_i - \mub^t}{\sqrt{\boldsymbol{\sigma}_{t}^{2} + \boldsymbol{\lambda}_{t}^{2}}} \right)^{2} - \boldsymbol{1} \right)\Bigg) \\
    \mathbf{w} &= \left[\kb_{\Xb\Xb} + \sigma_{noise}^{2}\mathbf{I}\right]^{-1} (\y  - \mb_{\Xb}) \odot \mathcal{N}\left(\Xb; \mub^{t}, \Lambda + \Sigmab^{t}\right)
    \end{split}
\end{equation}
\end{prop}
\begin{proof}
We follow the argument given in the proof of Proposition 16 in \cite{ollivier2017information}.
\end{proof}

\section{Experimental Details}

\subsection{Test Functions} \label{appendix:test_function}

We chose an initial distribution $\nu_{\theta} \sim \mathcal{N}(-\mathbf{1}, \mathbf{I})$. The synthetic test functions used are the following:

\textbf{Ackley}. The Ackley test function has analytical form: $f(x) = -A \exp(-B \sqrt{\frac{1}{d}\sum_{i=1}^d x_i^2}) - \exp(1/d \sum_{i=1}^d \cos(c x_i)) + A + exp(1)$. We chose the default parameters $d=2$ and $A = 20$, $B = 0.2$ and $C = 2\pi$ with optimum point at $\x = [0, \cdots, 0]$.

\textbf{Levy}. The Levy test function has analytical form: $f(x) = \sin^2(\pi w_1) + \sum_{i=1}^{d-1} (w_i-1)^2 (1 + 10 \sin^2(\pi w_i + 1)) + (w_d - 1)^2 (1 + \sin^2(2 \pi w_d))$. We chose the default parameters $d=2$ and $w_i = 1 + 0.25(\x_i - 1)$ with optimum point at $\x = [1, \cdots, 1]$.

\textbf{Styblinski Tang}. The Styblinski Tang test function has analytical form: $f(\x) = 0.5 * \sum_{i=1}^d (x_i^4 - 16x_i^2 + 5 x_i)$. We chose the default parameters $d=2$ with optimum point at $\x = [-2.903534, \cdots, -2.903534]$.

\textbf{Rastrigin}. The Rastrigin test function has analytical form: $f(x) = Ad + \sum_{i=1}^d (x_i^2 - A \cos{2\pi x_i})$. We chose the default parameters $d=2$ and $A = 10$ with optimum point at $\x = [0, \cdots, 0]$.

\textbf{Branin}. The Branin test function has analytical form: $f(x) = (x_2 - (b - 0.1 (1 - x_3))x_1^2 + c x_1 - r)^2 + 10(1-t) cos(x_1) + 10$. We chose the default parameters $d=2$ and $b = 5.1 / (4 \pi^2)$, $B = 0.2$ and $c = 5 / \pi$ $r = 6$, $t = 1 / (8 \pi)$ with infinitely many optimum.

\textbf{Griewank}. The Griewank test function has analytical form: $f(x) = \sum_{i=1}^d x_{i}^{2} / 4000 - \prod_{i=1}^d \cos(x_i / \sqrt{i}) + 1$ with optimum point at $\x = [0, \cdots, 0]$.

\textbf{Shekel}. The Shekel test function has analytical form: $f(x) = -\sum_{i=1}^{10} (\sum_{j=1}^4 (x_j - A_{ji})^2 + C_i)^{-1}$ with default parameter and with optimum point at $\x = [4, \cdots, 4]$.

\textbf{Three-Hump Camel Function}. The Three-Hump Camel Function test function has analytical form: $f(x) = 2x_{1}^{2} - 1.05 x_{1}^{4} + \x_{1}^{6}/6 + \x_{1}\x_{2} + \x_{2}^{2}$ with optimum point at $\x = [0, \cdots, 0]$.

\subsection{UCI Experiments} \label{appendix:uci}
For the UCI experiment we turned regression problems into an optimisation problem by training a surrogate model, namely a SVM model with RBF kernal model, that we optimise directly. We fit a multivariate distribution on the data and standardise them before fitting our surrogate model. As a result, we choose an initial distribution $\nu_{\theta} \sim \mathcal{N}(\mathbf{0}, \mathbf{I})$.

\textbf{Heart Disease}. This dataset consists of $d = 14$ features of patients and aims at predicting the probability of a heart disease being present in a given patient.

\textbf{Wine Quality}. This dataset predicts the quality of a certain wine given $d = 11$ features describing its chemical properties.

\textbf{Breast Cancer Wisconsin (Prognostic)}. This dataset predicts the probability for a patient to be diagnosed with breast cancer given $d = 30$ features that are computed from a digitised image of a fine needle aspirate (FNA) of a breast mass.  They describe characteristics of the cell nuclei present in the image.

\textbf{Fertility}. This dataset predicts normal or altered fertility from semen samples provided by individuals with $d = 11$ features relating to their health situation (e.g smoker, drinking alcohol ect...).

\textbf{Concrete Compressive Strength}. This dataset predicts the compressive strength of different concrete samples given $d = 9$ features related to its composition.

\textbf{Combined Cycle Power Plant}. This dataset predicts the energy output of a power plant given $d = 4$ internal characteristics: Temperature (T), Ambient Pressure (AP), Relative Humidity (RH) and Exhaust Vacuum (V).

\textbf{Appliances Energy Prediction}. This dataset predicts the energy use of appliances given $d = 28$ building characteristics gathered through sensors.

\textbf{Superconductivty Data}. This dataset predicts the critical temperature of different conductors given $d = 81$ features.

\textbf{Airfoil Self-Noise}. This dataset predicts the scaled sound pressure level, in decibels, of two and three-dimensional airfoil blade sections in an anechoic wind tunnel given $d = 5$ experimental parameters.

\textbf{Iranian Churn}. This dataset predicts whether a customer of a telecom company will end its relationship with the company given $d = 13$ characteristics, including call failures, complains, subscription length ect...

\subsection{Latent Space Optimisation} \label{appendix:lso}
In this task, we optimise a variable defined in the latent space of a pre-trained generative model in order to generate decoded samples which match a certain chosen class.

\textbf{Score Function}. The objective function was defined as follow: $f(\x) = \sum_{i = 1}^{d} w_i p_i$ where $\mathbf{p} = D(\x)$, with $D(\x)$ the selected decoder which output probability class and $w_i = \exp\left[-2\left(y_i - c\right)^{2}\right]$. This objective is maximised when the decoded data has a probability of being classified as $c$ is one.

\textbf{MNIST}. For MNIST dataset, we selected a VAE for the generative and a CNN for the classifier. Both architecture and pretrained weights can be found at $\href{https://botorch.org/tutorials/vae_mnist}{https://botorch.org/tutorials/vae_mnist}$. We chose two labels, one corresponding to the number two, the other to number five.

\textbf{CIFAR10}. For CIFAR10 dataset, we selected a GAN for the generative and a CNN for the classifier. Both architecture and pretrained weights can be found at \href{https://github.com/csinva/gan-vae-pretrained-pytorch/tree/master}{https://github.com/csinva/gan-vae-pretrained-pytorch/tree/master}. We chose two labels, one corresponding to the class frog, the other to the class horse.
\subsection{Hyperparameter Tuning}\label{appendix:hpo}
In this task we selected one hyperparameter tuning problem from the Profet suite, namely FCNet, which is generated via a meta-model built using a feed-forward neural network as a classification model and trained on 16 OpenML tasks. We also introduce a hyperparameter tuning task, GPmll consisting in optimising the hyperparameters of a Gaussian Process model fitting some data.

\textbf{FCNet} In this task, the objective function has six input parameters, corresponding to feed-forward hyperparameters learning rate, batch size, number of unit at layer one, probability of dropout at layer one, number of unit at layer two, probability of dropout at layer two. These parameter are transformed to be in the range $[0, 1]$. We considered an initial distribution $\nu_{\theta} \sim \mathcal{N}(\mathbf{0}, \mathbf{I})$ on a range [-3, 3] corresponding to three standard deviation that we scaled to fit the range $[0, 1]$.

\textbf{GPmll} In this task, the objective function has four input parameters, corresponding to the hyperparameters of a GP with constant mean and RBF kernel, namely the mean constant, likelihood noise, outputscale and the lengthscale. We took the default parameter set out in GPytorch on which we added a one standard deviation normal distribution. The marginal lilelihood is computed through fitting the GP to $100$ 1D points uniformly dispersed on the range $[0,1]$ and targets computed via $f(x) = \sin{2\pi x} + \epsilon$, $\epsilon \sim \mathcal{N}(0, 0.2)$.

\subsection{Locomotion Tasks}\label{appendix:rl}

\textbf{Cartpole}
This control task consists in maintaining a pole attached to a cart, which moves along a frictionless track. The pendulum is placed upright on the cart and the goal is to balance the pole by applying forces in the left and right direction on the cart. The observation space is four dimensional and consists in the Cart Position, Cart Velocity, Pole Angle and Pole Angular Velocity, and the action space consists in pushing the cart to the left or to the right. We optimise a linear policy of dimension $4$.

\textbf{Swimmer}
This control task consists in  moving swimmers as quickly as possible to the right. The swimmers are made up of three or more segments (called 'links'), connected by one fewer articulation points (referred to as 'rotors'). Each rotor links exactly two segments, forming a linear chain. The swimmer is placed in a two-dimensional pool and always starts from a fixed position (with some slight variation drawn from a uniform distribution). Applying torque to the rotors, leveraging fluid friction allows the swimmers to be propelled forward. This tasks has an action space of dimension $2$ (torque applied on first or second rotor) and an observation space of dimension $8$ corresponding to angles, velocity and angle velocities on various parts of the swimmers. We optimise a linear policy of dimension $16$.

In both tasks, we considered an initial distribution $\nu_{\theta} \sim \mathcal{N}(\mathbf{0}, \mathbf{3I})$ and bounds within $[-\mathbf{5}, \mathbf{5}]^{D}$ for our optimisation problem.

\section{Time Complexity Analysis}
We measured the wall-clock time that elapsed during the first $20$ iterations of the baselines algorithms that we presented, and for all the test function tasks. The results are shown in table \ref{table:time} where the time is given in seconds. As expected, the implementation of the natural evolutionary search algorithms is faster than their probabilistic counterparts. ProbNES is also significantly faster than vanilla BO and $\pi$-BO. While ProbNES is slower than ES algorithms, they are still more effective, as we demonstrated, in the scenario where we optimise a black box function which is potentially expensive to evaluate.

\begin{table}[h!]
\begin{center}
\begin{tabular}{|c c c c c c c c c|} \hline 

 baselines & Ackley & Rastrigin & ThreeHumpCamel & StyblinskiTang & Shekel & Levy & Griewank & Branin \\ [0.5ex] 
 \hline\hline
 Random & 0.0036 & 0.0030 & 0.0037 & 0.0030 & 0.0121 & 0.0054 & 0.0039 & 0.0036\\ 
 \hline
 CMAES & 0.0456 & 0.0407 & 0.0480 & 0.0413 & 0.1128 & 0.0690 & 0.0471 & 0.0512 \\
 \hline
 SNES & 0.0385 & 0.0353 & 0.0415 & 0.0344 & 0.1059 & 0.0624 & 0.0399 & 0.0444\\
 \hline
 XNES & 0.0556 & 0.0516 & 0.0589 & 0.0518 & 0.1241 & 0.0794 & 0.0569 & 0.0614\\
 \hline
 Prob-CMAES & 3.9436 & 2.8367 & 6.6701 & 6.4348 & 3.4978 & 5.8580 & 6.4230 & 8.4378\\
 \hline
  Prob-SNES & 4.6552 & 2.3915 & 6.2537 & 7.1182 & 3.1386 & 5.9210 & 8.4400 & 8.6828 \\
 \hline
  Prob-XNES & 3.5111 & 2.7448 & 5.2660 & 5.2365 & 3.4643 & 5.0714 & 5.0064 & 7.9972 \\
 \hline
 $\pi$-BO & 33.6572 & 30.3929 & 39.9827 & 33.9565 & 42.0564 & 35.0270 & 42.7260 & 34.2268 \\
 \hline
 Vanilla BO & 8.4744 & 11.1245 & 22.0442 & 11.7501 & 11.6407 & 19.3403 & 13.8111 & 14.9026 \\
 \hline
\end{tabular}
\caption{Wall-clock-time (s) for the first $20$ iterations of the baselines algorithm on test function tasks.}
\label{table:time}
\end{center}
\end{table}

\section{Ablation Study} \label{section:abl}
We studied the influence of certain problem and model parameter to the relative performance of ProbNES with respect to their NES counterparts, or how ProbNES performance are robust to such changes in the parameters. To do this, we ran the considered model to Ackley, Levy, and StyblinskiTang. The reason for this choice is the possibility to vary the input dimensionality of these synthetic functions.

\textbf{Influence of Dimension}.
\begin{figure}[h]
    \centering
\includegraphics[width=\textwidth]{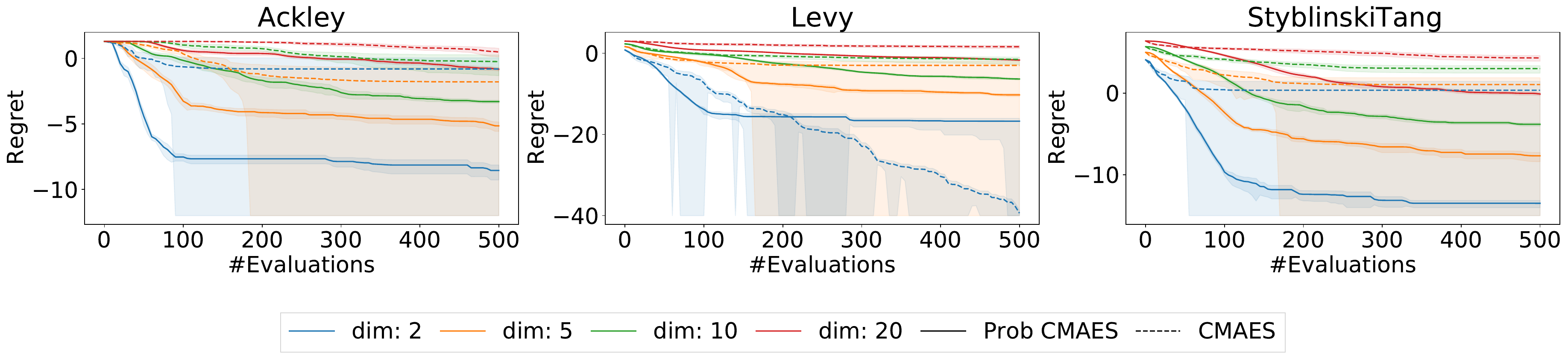}
    \caption{Influence of function dimension}
    \label{fig:abl_dim}
\end{figure}
We studied the influence of input dimension to the relative performance of ProbNES with respect to their NES counterparts. Figure \ref{fig:abl_dim} shows the log-regret of Prob CMAES and CMAES with dimension $2$, $5$, $10$ and $20$. Apart for the Levy function in dimension $2$, Prob CMAES consistently outperforms CMAES.

\textbf{Influence of the Acquisition Function}.
\begin{figure}[h]
    \centering
\includegraphics[width=\textwidth]{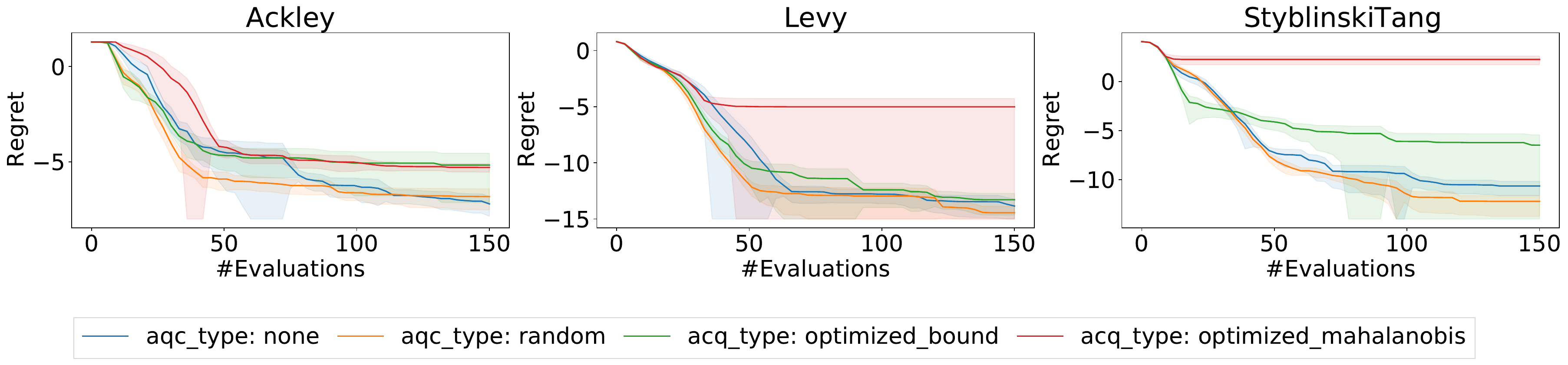}
    \caption{Influence acquisition function}
    \label{fig:abl_acq}
\end{figure}
We studied the influence of the acquisition function described in \ref{subsection:techimprov} to the performance of Prob XNES in Figure \ref{fig:abl_acq} which shows the log-regret. From this plot we deduce two things, the first is the comparison of "none" which correspond to random samples from the search distribution and "random", which corresponds to selecting the best batch of candidates that maximises the acquisition function, when these candidates are sample in large number (1000 batches of candidates). We can see that "random" outperforms "none" consistently, this proves that the acquisition is beneficial to the performance of the algorithm. Second, "random" also outerform "optimised bound" and "optimised mahalanobis", where the candidates are directly sampled through bounding boxes. This shows applying BQ directly in a bounding box tends to overexplore and leads to worse performance when it is not restricted to the search distribution. 

\textbf{Influence of Mahalanobis Parameter}.
\begin{figure}[h]
    \centering
\includegraphics[width=\textwidth]{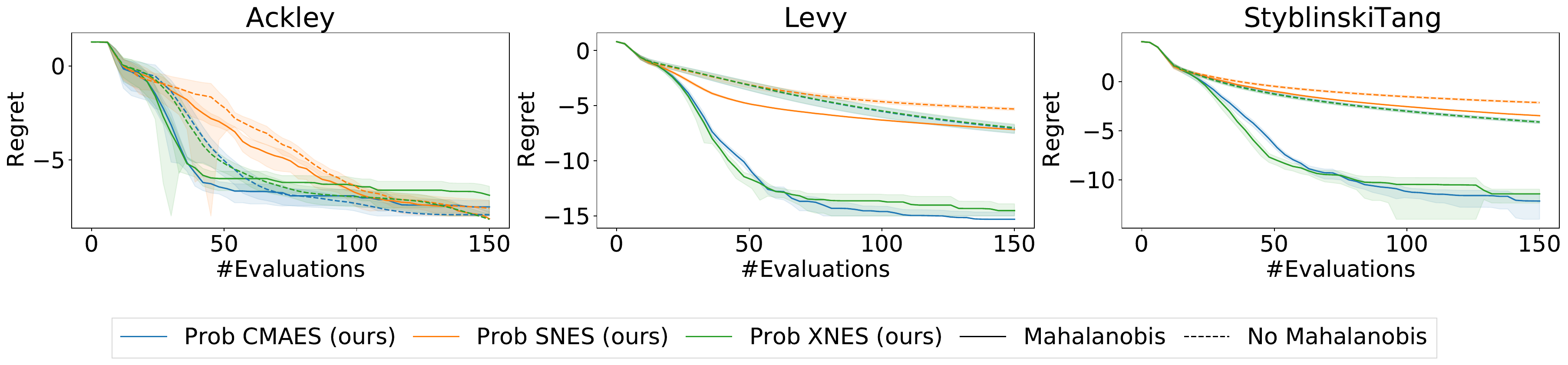}
    \caption{Influence Mahalanobis parameter}
    \label{fig:abl_mal}
\end{figure}
We studied the influence of introducing active and passive datapoints, as described in \ref{subsection:techimprov}, to the performance of Prob NES algorithms in Figure \ref{fig:abl_mal} which shows the log-regret. We can observe introducing this notion significantly and consistently improves performances, which also shows the GP tends to overfit in these synthetic functions.
\begin{figure}[h]
    \centering
\includegraphics[width=\textwidth]{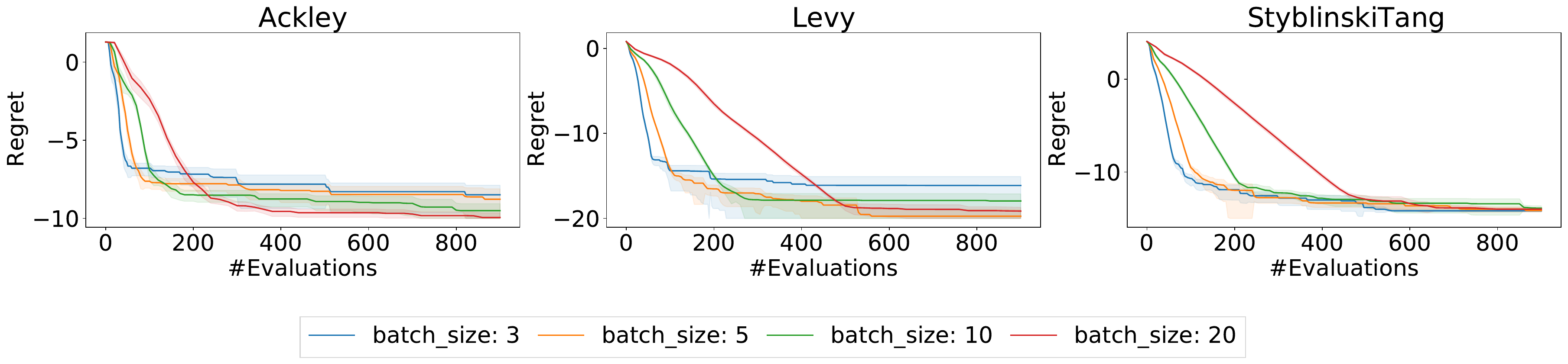}
    \caption{Influence batch size}
    \label{fig:abl_batch}
\end{figure}
We studied the influence of introducing active and passive datapoints, as described in \ref{subsection:techimprov}, on the performance of Prob NES algorithms in Figure \ref{fig:abl_mal} which shows the log-regret. We can observe introducing this notion significantly and consistently improves performances, which also shows the GP tends to overfit in these synthetic functions.

\textbf{Influence of Batch Size}.
We studied the influence of the batch size on the performance of Prob CMAES in Figure \ref{fig:abl_batch}. We can observe a higher batch size leads to better global performance but less sample efficiency.

\begin{figure}[h]
    \centering
\includegraphics[width=\textwidth]{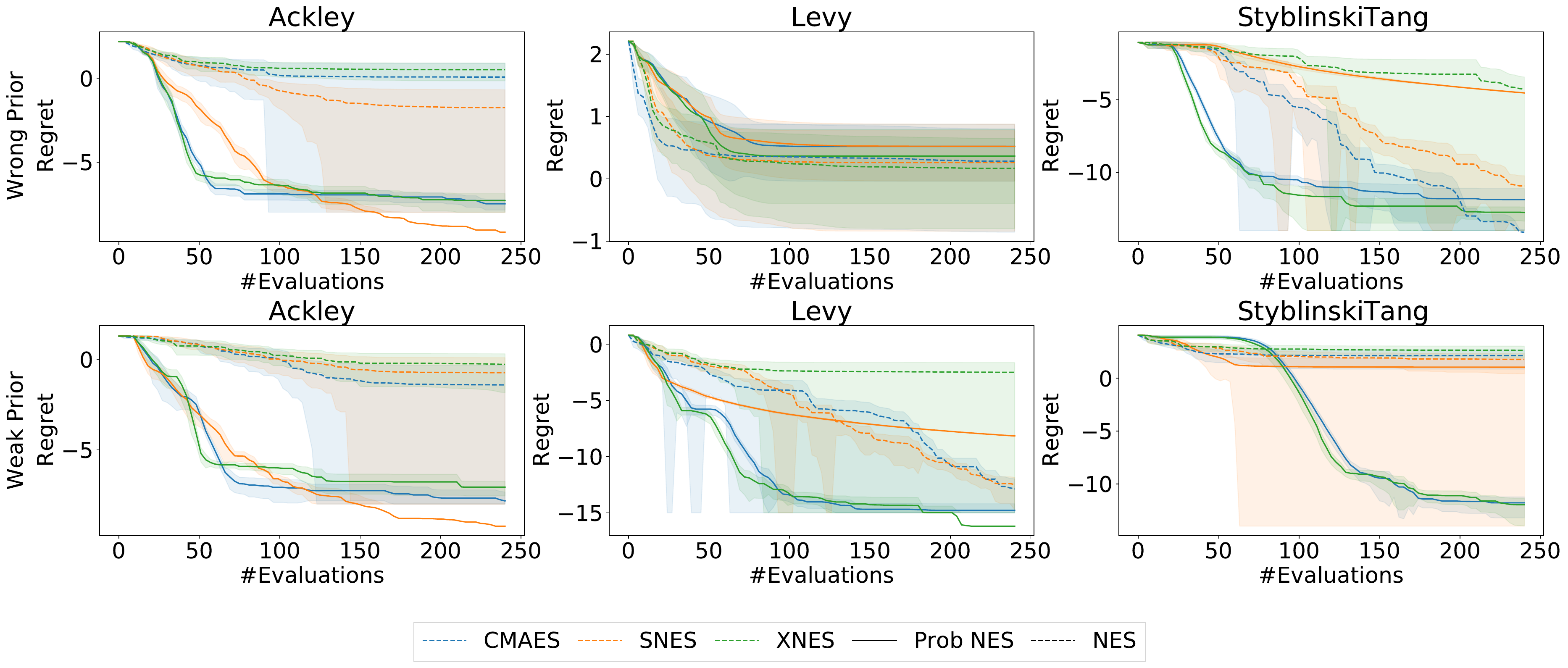}
    \caption{Influence prior strength}
    \label{fig:abl_prior}
\end{figure}

\textbf{Influence of Prior Strength}.
We studied the influence of the initial search distribution on the performance of Prob NES algorithms in Figure \ref{fig:abl_prior}. We define wrong prior as $\nu_{\theta} \sim \mathcal{N}(-\mathbf{3}, \mathbf{I})$ and weak prior $\nu_{\theta} \sim \mathcal{N}(-\mathbf{1}, 4\mathbf{I})$.

\textbf{Influence of Expected/Sampled Gradient}.
We studied the influence of the choice of gradient estimation, either expected or sampled, as described in \ref{subsection:PNES}. We can observe than choosing the expected or sampled gradient as very little influence, except in the experience with Ackley where the expected gradient consistently outperform the sampled one on all Prob NES algorithms.
\begin{figure}[h]
    \centering
\includegraphics[width=\textwidth]{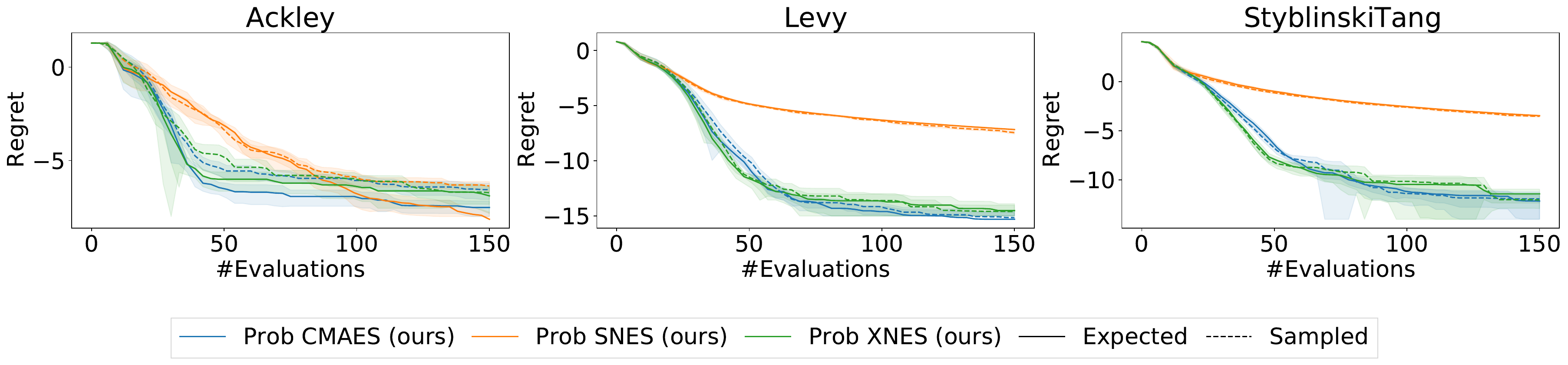}
    \caption{Influence expected or sampled gradient}
    \label{fig:abl_sampled}
\end{figure}

\end{document}